\documentclass[a4paper,11pt,abstract=on]{scrartcl}
\usepackage[utf8]{inputenc}
\usepackage[english]{babel} 
\usepackage[paper=a4paper,left=25mm,right=25mm,top=25mm,bottom=30mm]{geometry}
\usepackage{booktabs}
\usepackage{amsfonts,amsmath,amssymb}
\usepackage{bbm}
\usepackage{graphicx}
\usepackage{authblk}
\usepackage{natbib}
\usepackage[breaklinks, colorlinks=true, citecolor=black, urlcolor=black, linkcolor=black]{hyperref}

\title{Neural networks for post-processing ensemble weather forecasts}

\author[1]{Stephan Rasp}
\author[2,3]{Sebastian Lerch}
\affil[1]{Meteorological Institute, Ludwig-Maximilians-Universität, Munich}
\affil[2]{Institute for Stochastics, Karlsruhe Institute of Technology}
\affil[3]{Heidelberg Institute for Theoretical Studies}

\date{\today}

\begin{document}

\maketitle

\begin{abstract}
\noindent Ensemble weather predictions require statistical post-processing of systematic errors to obtain reliable and accurate probabilistic forecasts. Traditionally, this is accomplished with distributional regression models in which the parameters of a predictive distribution are estimated from a training period. We propose a flexible alternative based on neural networks that can incorporate nonlinear relationships between arbitrary predictor variables and forecast distribution parameters that are automatically learned in a data-driven way rather than requiring pre-specified link functions. In a case study of 2-meter temperature forecasts at surface stations in Germany, the neural network approach significantly outperforms benchmark post-processing methods while being computationally more affordable. Key components to this improvement are the use of auxiliary predictor variables and station-specific information with the help of embeddings. Furthermore, the trained neural network can be used to gain insight into the importance of meteorological variables thereby challenging the notion of neural networks as uninterpretable black boxes. Our approach can easily be extended to other statistical post-processing and forecasting problems. We anticipate that recent advances in deep learning combined with the ever-increasing amounts of model and observation data will transform the post-processing of numerical weather forecasts in the coming decade. 
\end{abstract}

\section{Introduction}

Numerical weather prediction based on physical models of the atmosphere has improved continuously since its inception more than four decades ago \citep{BauerEtAl2015}. In particular, the emergence of ensemble forecasts --- simulations with varying initial conditions and/or model physics --- added another dimension by quantifying the flow-dependent uncertainty. Yet despite these advances the raw forecasts continue to exhibit systematic errors which need to be corrected using statistical post-processing methods \citep{HemriEtAl2014}. Considering of the ever-increasing social and economical value of numerical weather prediction --- for example in the renewable energy industry --- producing accurate and calibrated probabilistic forecasts is an urgent challenge.

Most post-processing methods correct systematic errors in the raw ensemble forecast by learning a function that relates the response variable of interest to predictors. From a machine learning perspective, post-processing can be viewed as a supervised learning task. For the purpose of this study we will consider post-processing in a narrower distributional regression framework where the aim is to model the conditional distribution of the weather variable of interest given a set of predictors. The two most prominent approaches for probabilistic forecasts, Bayesian model averaging  \citep[BMA;][]{RafteryEtAl2005} and non-homogeneous regression, also referred to as ensemble model output statistics \citep[EMOS;][]{GneitingEtAl2005}, rely on parametric forecast distributions. This means one has to specify a predictive distribution and estimate its parameters, for example the mean and the standard deviation in the case of a Gaussian distribution. In the EMOS framework the distribution parameters are connected to summary statistics of the ensemble predictions through suitable link functions which are estimated by minimizing a probabilistic loss function over a training dataset. Including additional predictors, such as forecasts of cloud cover or humidity, is not straightforward in this framework and requires elaborate approaches to avoid overfitting \citep{MessnerEtAl2016}, a term that describes the inability of a model to generalize to data outside the training dataset. We propose an alternative approach based on modern machine learning methods, which is capable of including arbitrary predictors and learns nonlinear dependencies in a data-driven way.

Much work over the past years has been spent on flexible machine learning techniques for statistical modeling and forecasting \citep{McGovernEtAl2017}. Random forests \citep{Breiman2001}, for instance, can model nonlinear relationships including arbitrary predictors while being robust to overfitting. They have been used for classification and prediction of precipitation \citep{GagneEtAl2014}, severe wind \citep{LagerquistEtAl2017} and hail \citep{GagneEtAl2017}. In a post-processing context, quantile regression forest models have been proposed by \citet{TaillardatEtAl2016}. 

Neural networks are a flexible and user-friendly machine learning algorithm that can model arbitrary nonlinear functions \citep{Nielsen2015}. They consist of several layers of interconnected nodes which are modulated with simple nonlinearities (Figure \ref{fig:nn_schematic}; Section~\ref{sec:network-models}). Over the past decade many fields, most notably computer vision and natural language processing \citep{LeCunEtAl2015}, but also biology, physics and chemistry \citep{AngermuellerEtAl2016,GohEtAl2017} have been transformed by neural networks. In the atmospheric sciences, neural networks have been used to detect extreme weather in climate datasets \citep{LiuEtAl2016} and parameterize sub-grid processes in global circulation models \citep{GentineEtAl2018}. Neural networks have also been used for forecasting solar irradiances \citep{WangEtAl2012, ChuEtAl2013} and damaging winds \citep{LagerquistEtAl2017}. However, the complexity of the neural networks used in these studies was limited.

Here, we demonstrate how neural networks can be used for probabilistic post-processing of ensemble forecasts in the distributional regression framework. The presented model architecture allows for the incorporation of various features that are relevant for correcting systematic deficiencies of  ensemble predictions, and to estimate the network parameters by optimizing the continuous ranked probability score --- a mathematically principled loss functions for probabilistic forecasts. Specifically, we explore a case study of 2-meter temperature forecasts at surface stations in Germany with data from 2007--2016. We compare different neural network configurations to benchmark post-processing methods for varying training period lengths.  We further use the trained neural networks to gain meteorological insight into the problem at hand. Our ultimate goal is to present an efficient, multi-purpose approach to statistical post-processing and probabilistic forecasting. To the best of our knowledge, this study is the first to tackle  ensemble post-processing using neural networks.

The remainder of the paper is structured as follows: Section~\ref{sec:data} describes the forecast and observation data as well as the notation used throughout the study. In Section~\ref{sec:benchmark-models} we describe the benchmark post-processing models, followed by a description of the neural network techniques in Section~\ref{sec:network-models}. The main results are presented in Section~\ref{sec:results}. In Section~\ref{sec:importance} we explore the relative importance of the predictor variables. A discussion of possible extensions follows in Section~\ref{sec:discussion} before a conclusion in Section~\ref{sec:conclusion}.

\texttt{Python} \citep{Python} and \texttt{R} \citep{R} code for reproducing the results is available at \url{https://github.com/slerch/ppnn}.

\section{Data and notation}
\label{sec:data}

\subsection{Forecast data}
For this study, we focus on 2-meter temperature forecasts at surface stations in Germany at a forecast lead time of 48\,h. The forecasts are taken from the Interactive Grand Global Ensemble (TIGGE) dataset\footnote{available at \url{http://apps.ecmwf.int/datasets/data/tigge/}, see \url{https://github.com/slerch/ppnn/tree/master/data_retrieval}} \citep{BougeaultEtAl2010}. In particular, we use the global European Centre for Medium-Range Weather Forecasts (ECMWF) 50-member ensemble forecasts initialized at 00\,UTC every day. The data in the TIGGE archive is upscaled to a $0.5^\circ \times 0.5^\circ$ grid which corresponds to a horizontal grid spacing of around 35/55\,km (zonal/meridional). For comparison with the station observations, the gridded data were bilinearly interpolated to the observation locations. In addition to the target variable, we retrieved several auxiliary predictor variables (Table \ref{tab:features}). These were chosen broadly based on meteorological intuition. For each variable, we reduced the 50-member ensemble to its mean and standard deviation. 

Ensemble predictions are available from 3 January 2007 to 31 December 2016 every day. To ensure our verification procedure mimics operational conditions we set aside the year 2016 as a validation set. For model estimation we use two training periods, 2007--2015 and 2015 only, to assess the importance of training sample size.

\begin{table}
	\caption{Abbreviations and description of all features. Detailed definitions are available at \url{https://software.ecmwf.int/wiki/display/TIGGE/Parameters}. \label{tab:features}}
	\centering
	\begin{tabular}{ll}
	\toprule
		\textbf{Feature} & \textbf{Description}              \\ 
	\midrule 
	\multicolumn{2}{l}{\textit{Ensemble predictions (mean and standard deviation)}} \\
	\midrule 
		t2m          & 2-meter temperature                   \\
		cape         & Convective available potential energy \\
		sp           & Surface pressure                      \\
		tcc          & Total cloud cover                     \\
		sshf         & Sensible heat flux                    \\
		slhf         & Latent heat flux                      \\
		u10          & 10-meter U-wind                       \\
		v10          & 10-meter V-wind                       \\
		d2m          & 2-meter dew point temperature         \\
		ssr          & Short wave radiation flux             \\
		str          & Long wave radiation flux              \\
		sm           & Soil moisture                         \\
		v\_pl500     & V-wind at 500 hPa                     \\
		u\_pl500     & U-wind at 500 hPa                     \\
		u\_pl850     & U-wind at 850 hPa                     \\
		v\_pl850     & V-wind at 850 hPa                     \\
		gh\_pl500    & Geopotential at 500 hPa               \\
		q\_pl850     & Specific humidity at 850 hPa 		 \\
	\midrule 
	\multicolumn{2}{l}{\textit{Station-specific information}} \\
	\midrule 
		station\_alt & Altitude of station                   \\
		orog         & Altitude of model grid point          \\
		station\_lat & Latitude of station                   \\
				station\_lon & Longitude of station          \\
		\bottomrule         
	\end{tabular}
\end{table}

\subsection{Observation data}

\begin{figure}
\centering 
\includegraphics[width=0.5\textwidth]{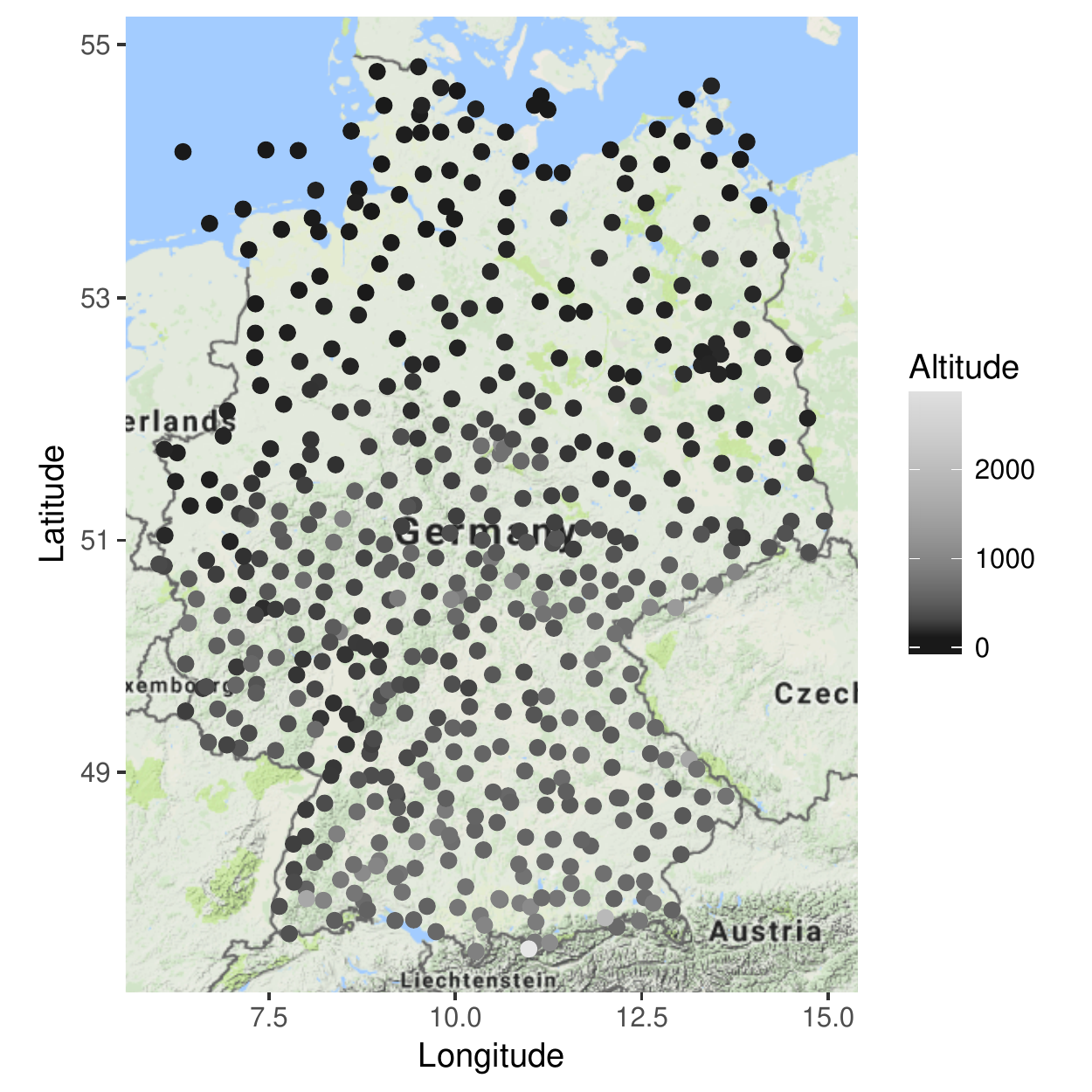}
\caption{Locations of DWD surface observation stations. The gray scale values of the points indicate the altitude in meters. \label{fig:map}}
\end{figure}

The forecasts are evaluated at 537 weather stations in Germany (see Figure \ref{fig:map}\footnote{All maps in this article were produced using the \texttt{R} package \texttt{ggmap} \citep{KahleWickham2013}.}). 2-meter temperature data are available from the Climate Data Center of the German weather service (Deutscher Wetterdienst, DWD)\footnote{available at \url{https://www.dwd.de/DE/klimaumwelt/cdc/cdc_node.html}}. Several stations have periods of missing data which are omitted from the analysis. During the evaluation period in calendar year 2016, observations are available at 499 stations. 

After removing missing observations, the 2016 validation set contains 182\,218 samples, the 2007--2015 training set contains 1\,626\,724 samples and the 2015 training set contains 180\,849 samples.

\subsection{Notation}

We now introduce the notation that is used throughout the rest of the paper. An observation of 2-meter temperature at station $s\in\{1,\dots,S\}$ and time $t\in\{1,\dots,T\}$ will be denoted by $y_{s,t}$. For each $s$ and $t$, the  50-member ECMWF ensemble forecast of variable $v$ is given by $ x_{s,t}^{v,1}, \dots,  x_{s,t}^{v,50},$ with mean value $x_{s,t}^{v,\text{mean}}$ and standard deviation $x_{s,t}^{v,\text{sd}}$. The mean values and standard deviations of all variables in the upper part of Table \ref{tab:features} are combined with station-specific features in the lower part, and aggregated into a vector of predictors $\boldsymbol{X}_{s,t} \in \mathbb{R}^p, p = 42$. Further, we write $\boldsymbol{X}_{s,t}^{\text{t2m}}$ to denote the vector of predictors that only contains mean value and standard deviations of the 2-meter temperature forecasts.

\section{Benchmark post-processing techniques}\label{sec:benchmark-models}

\subsection{Ensemble model output statistics}

In the general EMOS framework proposed by \citet{GneitingEtAl2005}, the conditional distribution of the weather variable of interest, $y_{s,t}$, given ensemble predictions $\boldsymbol{X}_{s,t}$, is modeled by a single parametric forecast distribution $ F_{{\boldsymbol{\theta}}_{s,t}}$ with parameters $\boldsymbol{\theta}_{s,t}\in\mathbb{R}^d$,
\begin{equation}
y_{s,t} | \boldsymbol{X}_{s,t} \sim F_{{\boldsymbol{\theta}}_{s,t}}.
\end{equation}
The parameters vary over space and time, and depend on the ensemble predictions $\boldsymbol{X}_{s,t}$ through suitable link functions $g:\mathbb{R}^p \rightarrow \mathbb{R}^d$,
\begin{equation}\label{eq:std-emos-link}
\boldsymbol{\theta}_{s,t} = g(\boldsymbol{X}_{s,t}).
\end{equation}

Here, we are interested in modeling the conditional distribution of temperature and follow \citet{GneitingEtAl2005} who introduce a model based on ensemble predictions of temperature, $\boldsymbol{X}_{s,t}^{\text{t2m}}$, only, where the forecast distribution is Gaussian with mean $\mu_{s,t}$ and standard deviation $\sigma_{s,t}$, i.e.,
\begin{equation*} 
y_{s,t} | \boldsymbol{X}_{s,t}^{\text{t2m}} \sim \mathcal{N}_{{(\mu_{s,t},\sigma_{s,t})}},
\end{equation*}
and where the link functions for mean and standard deviation are affine functions of the ensemble mean and standard deviation, respectively,
\begin{align}
(\mu_{s,t}, \sigma_{s,t}) &= g\left(\boldsymbol{X}_{s,t}^{\text{t2m}}\right) \nonumber \\ 
& = \left(a_{s,t} + b_{s,t}\,x_{s,t}^{\text{t2m},\text{mean}}, c_{s,t} + d_{s,t}\,x_{s,t}^{\text{t2m},\text{sd}}\right).\label{eq:std-emos}
\end{align}
Over the past decade, the EMOS framework has been extended from temperature to other weather variables including wind speed \citep{ThorarinsdottirGneiting2010, LerchThorarinsdottir2013, BaranLerch2015, ScheuererMoeller2015} and precipitation \citep{MessnerEtAl2014,Scheuerer2014, ScheuererHamill2015}.

The model parameters (or EMOS coefficients) $\kappa_{s,t} = (a_{s,t}, b_{s,t}, c_{s,t}, d_{s,t})$ are estimated by minimizing the mean continuous ranked probability score (CRPS) as function of the parameters over a training set. The CRPS is an example of a proper scoring rule, i.e., a mathematically principled loss function for distribution forecasts, and is a standard choice in meteorological applications. Details on the mathematical background of proper scoring rules and their use for model estimation are provided in Appendix \ref{sec:appendix-evaluation}.

Training sets are often considered to be comprised of the most recent days only. However, as we did not find substantial differences in predictive performance (see Section \ref{sec:results}), we estimate the coefficients over a fixed training set, they thus do not vary over time and we denote them by $\kappa_{s}$. Estimation is usually either performed locally, i.e., considering only forecast cases from the station of interest, or globally by pooling together forecasts and observations from all stations. We refer to the corresponding EMOS models by EMOS-loc and EMOS-gl, respectively. The parameters $\kappa$ of the global model do not depend on the station $s$, and are thus unable to correct location-specific deficiencies of the ensemble forecasts. Alternative approaches where training sets are selected based on similarities of weather situations or observation station characteristics were proposed by \citet{JunkEtAl2015} and \citet{LerchBaran2017}. EMOS-gl and EMOS-loc are implemented in \texttt{R} with the help of the \texttt{scoringRules} package \citep{JordanEtAl2017}.

\subsection{Boosting for predictor selection in EMOS models}

Extending the EMOS framework to allow for including additional predictor variables is non-trivial as the increased number of parameters can result in overfitting. \citet{MessnerEtAl2017} proposed a boosting algorithm for this purpose. In this approach components of the link function $g$ in \eqref{eq:std-emos-link} are chosen to be an affine function for the mean $\mu_{s,t}$ and an exponential transformation of an affine function for the standard deviation $\sigma_{s,t}$, 
\begin{align}
(\mu_{s,t}, \sigma_{s,t}) &= g(\boldsymbol{X}_{s,t}) \nonumber \\ 
	& = \left((1,\boldsymbol{X}_{s,t})^\text{T}\boldsymbol{\beta}_{s,t}, \ \exp( (1,\boldsymbol{X}_{s,t})^\text{T}\boldsymbol{\gamma}_{s,t} ) \right). \label{eq:bst}
\end{align}
Here, $\boldsymbol{\beta}_{s,t}\in\mathbb{R}^{p+1}$ and $\boldsymbol{\gamma}_{s,t}\in\mathbb{R}^{p+1}$ denote coefficient vectors corresponding to the vector of predictors $\boldsymbol{X}_{s,t}$ extended by a constant. As for the standard EMOS models, the coefficient vectors are estimated over fixed training periods and thus do not depend on $t$, we suppress the index in the following.

The boosting algorithm proceeds iteratively by updating the coefficient of the predictor that improves the current model fit most. As the coefficient vectors are initialized as $\boldsymbol{\beta}_{s} = \boldsymbol{\gamma}_{s} = \boldsymbol{0}$, only the most important variables will have non-zero coefficients if the algorithm is stopped before convergence. The contributions of the different predictors are assessed by computing average correlations to partial derivatives of the loss function with respect to $\mu_{s,t}$ and $\sigma_{s,t}$ over the training set. If the current model fit is improved, the coefficient vectors are updated by a pre-defined step size into the direction of steepest descent of linear approximations of the gradients. 

We denote local EMOS models with an additional boosting step by EMOS-loc-bst. The tuning parameters of the algorithm were chosen by fitting models for a variety of choices and picking the configuration with the best out-of-sample predictions (Table \ref{tab:hyper-parameters-benchmark}) based on implementations in the \texttt{R} package \texttt{crch} \citep{MessnerEtAl2016}. Note, however, that the results are not very sensitive to the exact choice of tuning parameters. For the local model considered here, the station-specific features in the lower part of Table \ref{tab:features} are not relevant and are excluded from $\boldsymbol{X}_{s,t}$. Boosting-based variants of global EMOS models have also been tested, but result in worse forecasts.

The boosting-based EMOS-loc-bst model differs from the standard EMOS models (EMOS-gl and EMOS-loc) in several aspects. First, the boosting step allows to include covariate information from predictor variables other than temperature forecasts. Second, the parameters are estimated by maximum likelihood estimation, i.e., by minimizing the mean logarithmic score by contrast to minimum CRPS estimation, see Appendix \ref{sec:appendix-evaluation} for details. Further, the affine link function for the standard deviation in \eqref{eq:std-emos} is replaced by an affine function for the logarithm of the standard deviation in \eqref{eq:bst}.

\subsection{Quantile regression forests}\label{sec:benchmark-models-qrf}

Parametric distributional regression models such as the EMOS methods described above require the choice of a suitable parametric family $F_{\boldsymbol{\theta}}$. While the conditional distribution of temperature can be well approximated by a Gaussian distribution, this poses a limitation for other weather variables such as wind speed or precipitation where the choice is less obvious \citep[see, e.g.,][]{BaranLerch2018}. 

Non-parametric distributional regression approaches provide alternatives that circumvent the choice of parametric family. For example, quantile regression approaches approximate the conditional distribution by a set of quantiles. In the context of post-processing ensemble forecasts, \citet{TaillardatEtAl2016} proposed a quantile regression forest (QRF) model based on the work of \citet{Meinshausen2006} that allows to include additional predictor variables. 

The QRF model is based on the idea of generating random forests from classification and regression trees \citep{BreimanEtAl1984}. These are binary decision trees obtained by iteratively splitting the training data into two groups according to some threshold for one of the predictors, chosen such that every split minimizes the sum of the variance of the response variable in each of the resulting groups. The splitting procedure is iterated until a stopping criterion is reached. The final groups (or terminal leaves) thus contain subsets of the training observations based on the predictor values, and out of sample forecasts at station $s$ and time $t$ can be obtained by proceeding through the decision tree according to the corresponding predictor values $\boldsymbol{X}_{s,t}$. Random forest models \citep{Breiman2001} increase stability of the predictions by averaging over many random decision trees generated by selecting a random subset of the predictors at each candidate split in conjunction with bagging, i.e., bootstrap aggregation of random subsamples of training sets. In the quantile regression forest approach, each tree provides an approximation of the distribution of the variable of interest given by the empirical cumulative distribution function (CDF) of the observation values in the terminal leaf associated with the current predictor values $\boldsymbol{X}_{s,t}$. Quantile forecasts can then be computed from the combined forecast distribution which is obtained by averaging over all tree-based empirical CDFs.

We implement a local version of the QRF model where separate models are estimated for each station based on training sets that only contain past forecasts and observations from that specific station. As discussed by \citet{TaillardatEtAl2016}, the predicted quantiles are necessarily restricted to the range of observed values in the training period by construction which may be disadvantageous in case of shorter training periods. However, global variants of the QRF model did not result in improved forecast performance even with only one year of training data, we will thus restrict attention to the local QRF model. The models are implemented using the \texttt{quantregForest} package \citep{quantregForest} for \texttt{R}. Tuning parameters are chosen as for the EMOS-loc-bst model (Table \ref{tab:hyper-parameters-benchmark}).

The QRF approach has recently been extended into several directions. \citet{AtheyEtAl2016} propose a generalized version of random forest-based quantile regression based on theoretical considerations (GRF) which has been tested but did not result in improved forecast performance. \citet{TaillardatEtAl2017} combine QRF (and GRF) models and parametric distributional regression by fitting a parametric CDF to the observations in the terminal leaves instead of using the empirical CDF. \citet{SchlosserEtAl2018} combine parametric distributional regression and random forests for parameter estimation in the framework of a generalized additive model for location, scale and shape.   

\section{Neural networks}\label{sec:network-models}

In this section we will give a brief introduction to neural networks. For a more detailed treatment the interested reader is referred to more comprehensive resources \citep[e.g.,][]{Nielsen2015, Goodfellow2016}. The network techniques are implemented using the \texttt{Python} libraries Keras \citep{Keras} and TensorFlow \citep{Tensorflow}.

\begin{figure}
\centering 
\includegraphics[width=\textwidth]{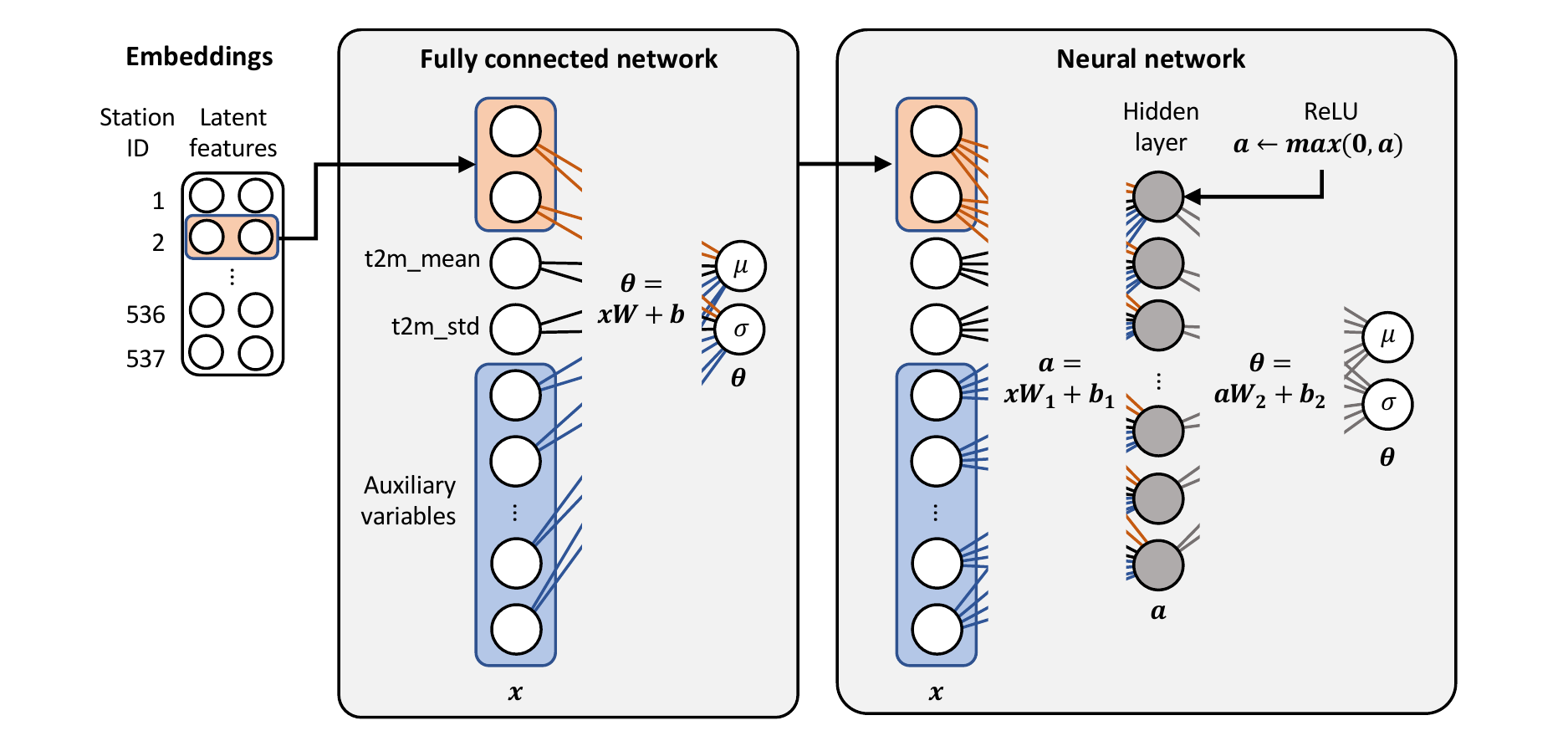}
\caption{Schematic of a fully-connected (left) and a neural network with one hidden layer (right). In both cases, data flows from left to right. Orange nodes and connections illustrate station embeddings, blue ones auxiliary input variables. Mathematical operations are to be understood as element-wise operations for vector objects.}
\label{fig:nn_schematic}
\end{figure}

Neural networks consist of several layers of nodes (Figure \ref{fig:nn_schematic}), each of which is a weighted sum of all nodes $j$ from the previous layer plus a bias term: 
\begin{equation}
\sum_j w_j z_j + b
\end{equation}
The first layer contains the input values, or features, while the last layer represents the output values, or targets. In the layers in-between, called hidden layers, each node value is passed through a nonlinear activation function. For this study, we use a rectified linear unit (ReLU): 
\begin{equation*}
\text{ReLU}(z) = \max(0, z).
\end{equation*}
This activation function allows the neural network to represent nonlinear functions. The weights and biases are optimized to reduce a loss function using stochastic gradient descent (SGD). Here we employ an SGD version called Adam \citep{KingmaBa2014}.

In this study we use networks without a hidden layer and with a single hidden layer (Figure \ref{fig:nn_schematic}). The former, which we will call fully-connected networks (FCN), model the outputs as a linear combination of the inputs. The latter, called neural networks (NN) here, are capable of representing nonlinear relationships. 
Introducing additional hidden layers to neural networks did not improve the predictions as additional model complexity increases the potential of overfitting.

\subsection{Neural networks for ensemble post-processing}

Neural networks can be applied to a range of problems, such as regression and classification. The main difference between those are the  contents and activation function of the output layer, and the loss function. Here we use the neural network for the distributional regression task of post-processing ensemble forecasts. Our output layer represents the distribution parameters $\mu_{s,t}$ and $\sigma_{s,t}$ of the Gaussian predictive distribution. No activation function is applied. The corresponding probabilistic forecast describes the conditional distribution of the observation $y_{s,t}$ given the predictors $\boldsymbol{X}_{s,t}$ as input features. As a loss function we use the closed form expression of the CRPS for a Gaussian distribution, see equation \eqref{eq:crps-normal}. This is a non-standard choice in the neural network literature (\citet{DIsantoPolsterer2018} is the only previous study to our knowledge) but  provides a mathematically principled choice for the distributional regression problem at hand (see Appendix \ref{sec:appendix-evaluation} for the mathematical background). Other probabilistic neural network approaches include quantile regression \citep{Taylor2000} and distribution-to-distribution regression \citep{KouEtAl2018}. 

The simplest network model is a fully connected model based on predictors $\boldsymbol{X}_{s,t}^{\text{t2m}}$, i.e., mean and standard deviation of ensemble predictions of temperature only (denoted by FCN). Apart from additional connections for the mean and standard deviation to ensemble standard deviation and mean, respectively, the FCN model is conceptually equivalent to EMOS-gl, but differs in the parameter estimation approaches. A neural network with a hidden layer for the $\boldsymbol{X}_{s,t}^{\text{t2m}}$-input did not show any improvements over the simple linear model suggesting that there is no nonlinear relationships to exploit. Additional information from auxiliary variables can be taken into account by considering the entire vector $\boldsymbol{X}_{s,t}$ of predictors as input features. The corresponding fully connected and neural network models are referred to as FCN-aux and NN-aux. 

\subsection{Station embeddings}

To enable the networks to learn station-specific information we use embeddings, a common technique in natural language processing and recommender systems. An embedding is a mapping from a discrete object, in our case the station ID,  to a vector of real numbers. The elements of this vector with length $n_\text{emb}$ are also referred to as latent features. In total, therefore, the embedding matrix has dimension $S \times n_\text{emb}$, where $S$ is the number of stations. The latent features are concatenated with the predictors, $\boldsymbol{X}_{s,t}^{\text{t2m}}$ or $\boldsymbol{X}_{s,t}$, and are updated along with the weights and biases during training. This allows the algorithm to learn a specific set of numbers for each station. Here we use $n_\text{emb} = 2$ as larger values did not improve the predictions.

The fully connected network with input features $\boldsymbol{X}_{s,t}^{\text{t2m}}$ and embeddings is abbreviated by FCN-emb. As with FCN, adding a hidden layer did not improve the results. Fully connected and neural networks with both, station embeddings and auxiliary inputs $\boldsymbol{X}_{s,t}$, are denoted by FCN-aux-emb and NN-aux-emb.

\subsection{Further network details}

Neural networks with a large number of parameters, i.e., weights and biases, can suffer from overfitting. One way to reduce overfitting is to stop training early. When to stop can be guessed by taking out a subset (20\%) from the training set (2007--2015 or 2015) and checking when the score on this separate dataset stops improving. This gives a good approximation when to stop training on the full training set without using the actual 2016 validation set during training.
Other common regularization techniques to prevent overfitting, such as dropout or weight decay (L2 regularization), were not successful in our case.

Finally, we train ensembles of ten neural networks with different random initial parameters for each configuration and average over the forecast distribution parameter estimates to obtain $\boldsymbol{\theta}_{s,t}$. For the  more complex network models this helps to stabilize the parameter estimates by reducing the variability due to random variations between model runs and slightly improves the forecasts.

\section{Results}\label{sec:results}

Tuning parameters for all benchmark and network models are listed in Tables \ref{tab:hyper-parameters-benchmark} and \ref{tab:hyper-parameters-network}. Details on the employed evaluation methods are provided in Appendix \ref{sec:appendix-evaluation}.  

\subsection{General results}

The CRPS values averaged over all stations and the entire 2016 validation period are summarized in Table \ref{tab:results}. For the 2015 training period, EMOS-gl gives a 13\% relative improvement compared to the raw ECMWF ensemble forecasts in terms of mean CRPS.\footnote{Note that standard EMOS models are often implemented with rolling training periods, i.e., parameter estimation is based on the $m$ preceding days. Such implementations have been tested, but only result in marginal improvements. For the longer training period from 2007--2015, we have also implemented EMOS models that are estimated based on a centered window $[d_0-m, d_0+m]$ around the current day $d_0$ from all previous years. For the local model, this results in a slightly improved mean CRPS of 0.88, but does not improve the forecasts of EMOS-gl.} As expected, FCN which mimics the design of EMOS-gl achieves a very similar score. Adding local station information in EMOS-loc and FCN-emb improves the global score by another 10\%. While EMOS-loc estimates a separate model for each station, FCN-emb can be seen as a global network-based implementation of EMOS-loc. Adding covariate information through auxiliary variables results in an improvement for the fully connected models similar to that of adding station information. Combining auxiliary variables and station embeddings in FCN-emb-aux improves the mean CRPS further to 0.88 but the effects do not stack linearly. Adding covariate information in EMOS models using boosting (EMOS-loc-bst) outperforms FCN-emb-aux by 3\%. Allowing for nonlinear interactions of station information and auxiliary variables using a neural network (NN-aux-emb) achieves the best results, improving the best benchmark technique (EMOS-loc-bst) by 3\% for a total improvement compared to the raw ensemble of 29\%. The QRF model is unable to compete with most of the post-processing models for the 2015 training period.

The relative scores and model rankings for the 2007--2015 training period closely match those of the 2015 period. For the linear models (EMOS-gl, EMOS-loc and all FCN) more data does not improve the score by much. For EMOS-loc-bst and the neural network models, however, the skill is increased by 4--5\%. This suggests that longer training periods are most efficiently exploited by more complex, nonlinear models. QRF improves the most, now being among the best models, which indicates a minimum data amount required for this method to work. This is likely due to the limitation of predicted quantiles to the range of observed values in the training data, see Section \ref{sec:benchmark-models-qrf}.

\begin{table}[t]
\centering
\caption{Mean CRPS for raw and post-processed ECMWF ensemble forecasts, averaged over all available observations in calendar year 2016. The lowest (i.e., best) values are marked in bold font.}
\label{tab:results}
\begin{tabular}{llcc}
\toprule
Model & Description & \multicolumn{2}{c}{Mean CRPS for} \\
 &	&\multicolumn{2}{c}{training period} \\[0.15cm]
 &	& 2015 & 2007--2015 \\
\midrule 
Raw ensemble & & 1.16 & 1.16 \\
\midrule
\multicolumn{4}{l}{\textit{Benchmark post-processing methods}} \\
\midrule
EMOS-gl & Global EMOS & 1.01 & 1.00 \\
EMOS-loc & Local EMOS & 0.90 & 0.90 \\
EMOS-loc-bst & Local EMOS with boosting & 0.85 & 0.80 \\
QRF & Local quantile regression forest & 0.95 & 0.81 \\
\midrule
\multicolumn{4}{l}{\textit{Neural network models}} \\
\midrule 
FCN & Fully connected network & 1.01 & 1.01 \\
FCN-aux & \dots with auxiliary predictors & 0.92 & 0.91 \\
FCN-emb & \dots with station embeddings & 0.91 & 0.91 \\
FCN-aux-emb & \dots with both of the above & 0.88 & 0.87 \\
NN-aux & 1-hidden layer neural network with auxiliary predictors & 0.90 & 0.86 \\
NN-aux-emb & \dots and station embeddings & \textbf{0.82} & \textbf{0.78} \\
\bottomrule
\end{tabular}
\end{table}

To assess calibration, verification rank and probability integral transform (PIT) histograms of raw and post-processed forecasts are shown in Figures \ref{fig:pits-2015} and \ref{fig:pits-2007-15}. The raw ensemble forecasts are under-dispersed as indicated by the U-shaped verification rank histogram, i.e., observations tend to fall outside the range of the ensemble too frequently. By contrast, all post-processed forecast distributions are substantially better calibrated and the corresponding PIT histograms show much smaller deviations from uniformity. All models show a slight overprediction of high temperatures and, with the exception of QRF, an underprediction of low values. The linear EMOS and FCN models as well as QRF are further slightly overdispersive as indicated by inverse U-shaped upper parts of the histogram.

\subsection{Station-by-station results}

\begin{figure}[p]
\centering
CRPSS relative to raw ensemble forecasts \\
\includegraphics[width = \textwidth]{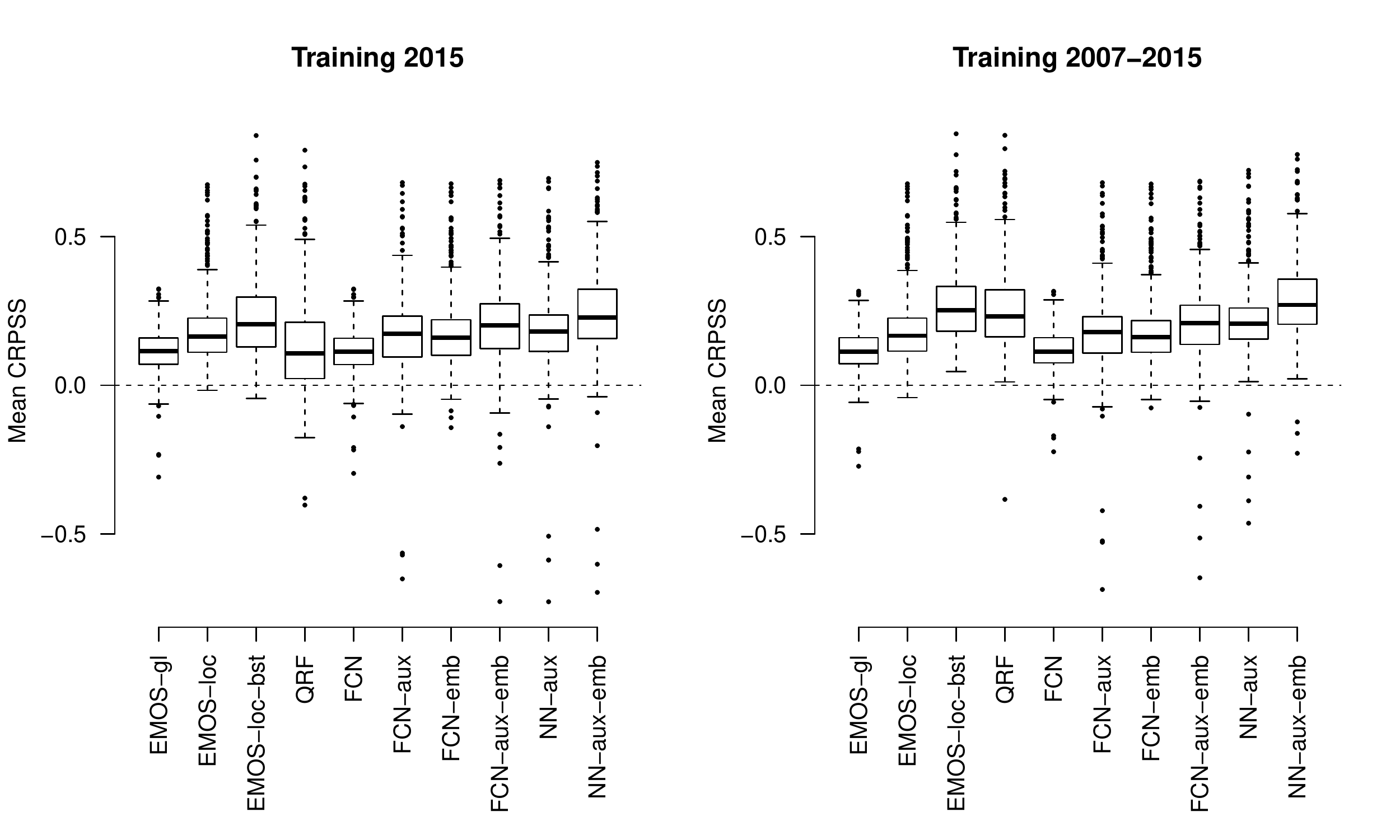} \\
\bigbreak 
CRPSS relative to EMOS-loc \\
\includegraphics[width = \textwidth]{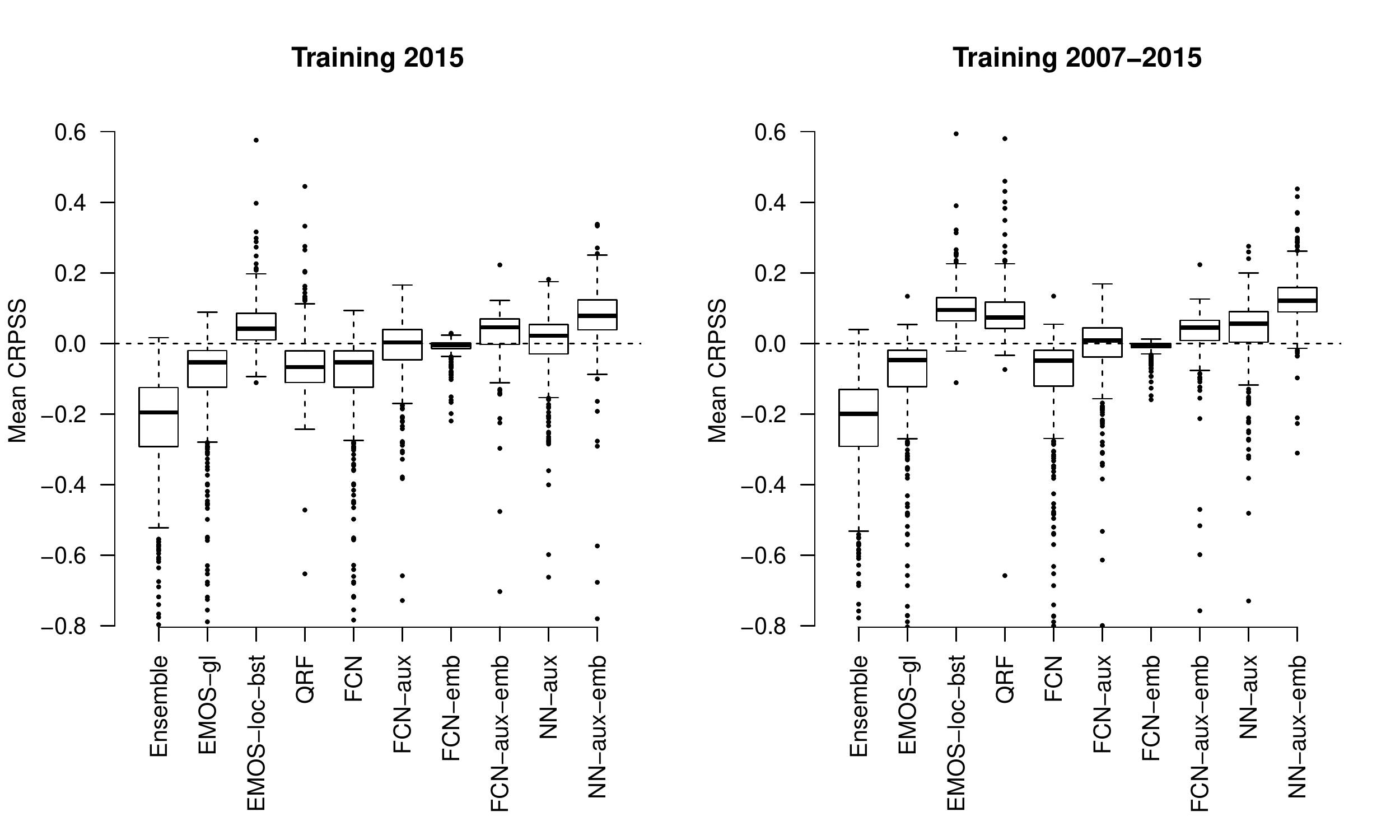} 
\caption{Boxplots of station-wise mean continuous ranked probability skill score of all post-processing models using the raw ensemble (top row) and EMOS-loc (bottom row) as reference forecast. A dot within each box represents the mean CRPSS at one of the observation stations. The CRPSS is computed so that positive values indicate an improvement of the model specified on the horizontal axis over the reference. Similar plots with different reference models are provided in Appendix \ref{sec:appendix-results-crpss}. \label{fig:crpss}}
\end{figure}

Figure~\ref{fig:crpss} shows the station-wise distribution of the continuous ranked probability skill score (CRPSS), which measures the probabilistic skill relative to a reference model. Positive values indicate an improvement over the reference. Compared to the raw ensemble, forecasts at most stations are improved by all post-processing methods with only a few negative outliers. Compared to EMOS-loc only FCN-aux-emb, the neural network models and EMOS-loc-bst show improvements at the majority of stations. Corresponding plots with the three best-performing models as reference experiments are provided in Figure \ref{fig:crpss-appendix}. It is interesting to note that the network models, with the exception of FCN and FCN-emb, have more outliers, particularly for negative values compared to the EMOS methods and QRF which have very few negative outliers. This might be due to a few stations with strongly location-specific error characteristics that the locally estimated benchmark models are better able capture. Training with data from 2007 to 2015 alleviates this somewhat. 

\begin{figure}
\centering 
\includegraphics[width = \textwidth]{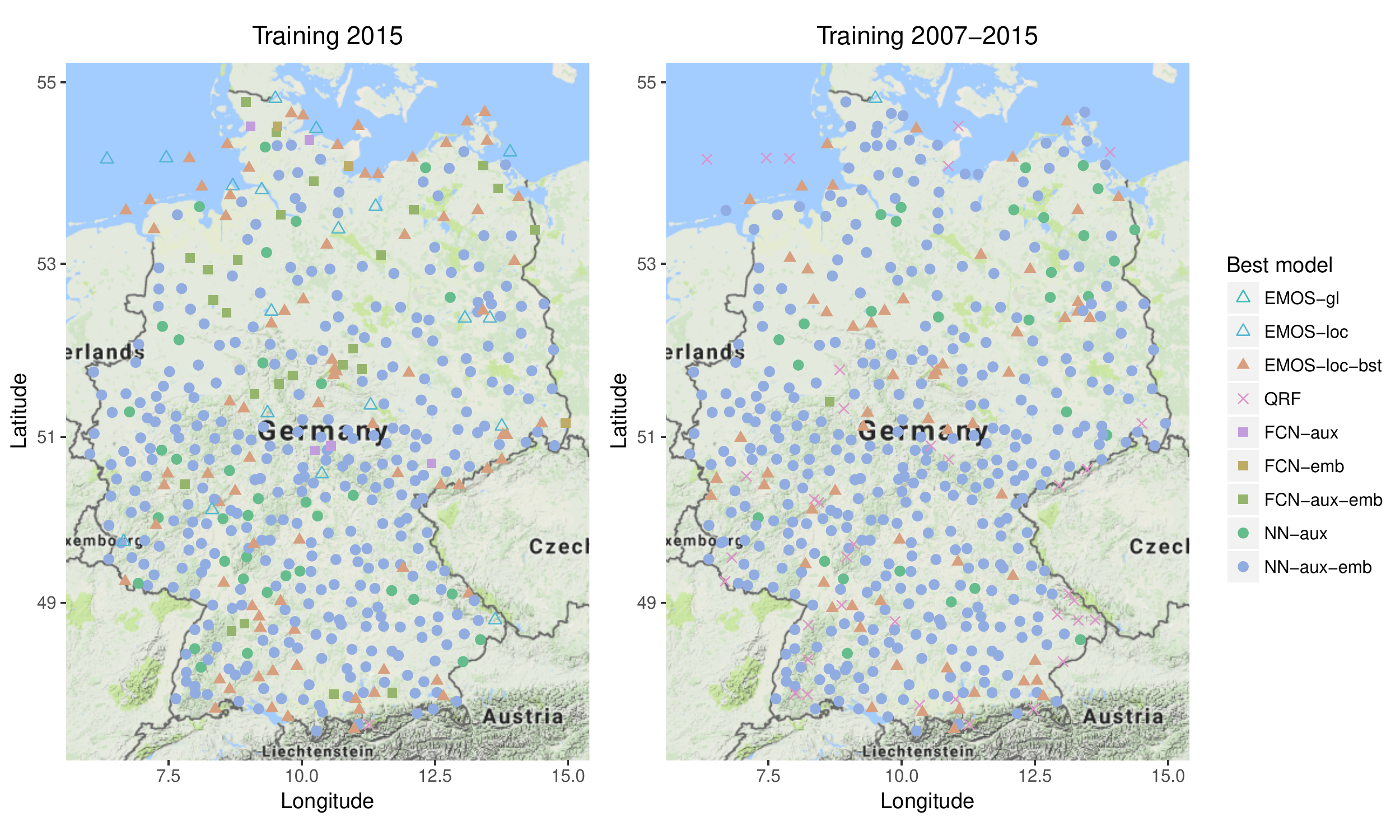} 
\caption{Observation station locations color-coded by the best performing model (in terms of mean CRPS over calendar year 2016) for models trained on data from 2015 (left), and on data from 2007--2015 (right). Point shapes indicate the type of model. \label{fig:map-bestmodel}}
\end{figure}

Figure \ref{fig:map-bestmodel} shows maps with the best-performing models in terms of mean CRPS for each station. For the majority of stations NN-aux-emb provides the best predictions. The variability of station-specific best models is greater for the 2015 training period compared to 2007--2015. The top three models for the 2015 period are NN-aux-emb (best at 65.9\% of stations), EMOS-loc-bst (16.0\%) and NN-aux (7.2\%), and for 2007--2015 NN-aux-emb (73.5\%), EMOS-loc-bst (12.4\%) and QRF (7.4\%). At coastal and offshore locations, particularly for the shorter training period, the benchmark methods tend to outperform the network methods. Ensemble forecast errors at these location likely have a strong location-specific component that might be easier to capture for the locally estimated EMOS and QRF methods.

Additionally, we evaluated the statistical significance of the differences between the competing post-processing methods using a combination of Diebold-Mariano tests \citep{DieboldMariano1995} and a \citet{BenjaminiHochberg1995} procedure to account for temporal and spatial dependencies of forecast errors. We thereby follow suggestions of \citet{Wilks2016}, mathematical details are deferred to Appendix \ref{sec:appendix-tests}. The results provided in Appendix \ref{sec:appendix-results-significance} generally indicate high ratios of stations with significant score differences in favor of the neural network models. Even when compared to the second best-performing model, EMOS-loc-bst,  NN-aux-emb is significantly better at 30\% of the stations and worse at only 2\% or less for both training periods. 

\subsection{Computational aspects}

While a direct comparison of computation times for the different methods is difficult, even the most complex network methods are a factor of two or more faster than EMOS-loc-bst. This includes creating an ensemble of ten different model realizations. QRF is by far the slowest method, being roughly ten times slower than EMOS-loc-bst. Complex neural networks benefit substantially from running on a graphics processing unit (GPU) compared to running on the core processing unit (CPU; roughly six times slower for NN-aux-emb). 
Neural network-ready GPUs are now widely available in many scientific computing environments or via cloud computing\footnote{For example, see \url{https://colab.research.google.com/}.} For more details on the computational methods and results see Appendix \ref{sec:appendix-results-computation}.

\section{Feature importance}
\label{sec:importance}

To assess the relative importance of all features we use a technique called permutation importance that was first described in the context of random forests \citep{Breiman2001}. We randomly shuffle each predictor/feature in the validation set one at a time and observe the increase in mean CRPS compared to the unpermuted features. While unable to capture colinearities between features this method does not require re-estimating the model with each individual feature omitted.

Consider a random permutation of station and time indices $\pi(s,t)$
and let $\boldsymbol{X}_{s,t}^{\text{perm}_v}$ denote the vector of predictors where variable $v$ is permuted according to $\pi$, i.e., a vector with $j$-th entry
\[
\boldsymbol{X}_{s,t}^{\text{perm}_v\,(j)} = 
\begin{cases}
\boldsymbol{X}^{(j)}_{s,t}, & j \neq v \\
\boldsymbol{X}^{(v)}_{\pi(s,t)} & j = v
\end{cases} \quad \text{for}\ j = 1,\dots,p.
\]
The importance of input feature $v$ is computed as mean CRPS difference
\[
\text{Importance}(v) = \frac{1}{ST} \sum_{s=1}^{S}\sum_{t=1}^{T} \left( \text{CRPS}(F|\boldsymbol{X}_{s,t}^{\text{perm}_v}, y_{s,t}) 
- \text{CRPS}(F|\boldsymbol{X}_{s,t}, y_{s,t}) \right),
\]
where we average over the entire evaluation set and $F|\boldsymbol{X}$ denotes the conditional forecast distribution given a vector of predictors.

\begin{figure}
	\centering
	\includegraphics[width=0.85\textwidth]{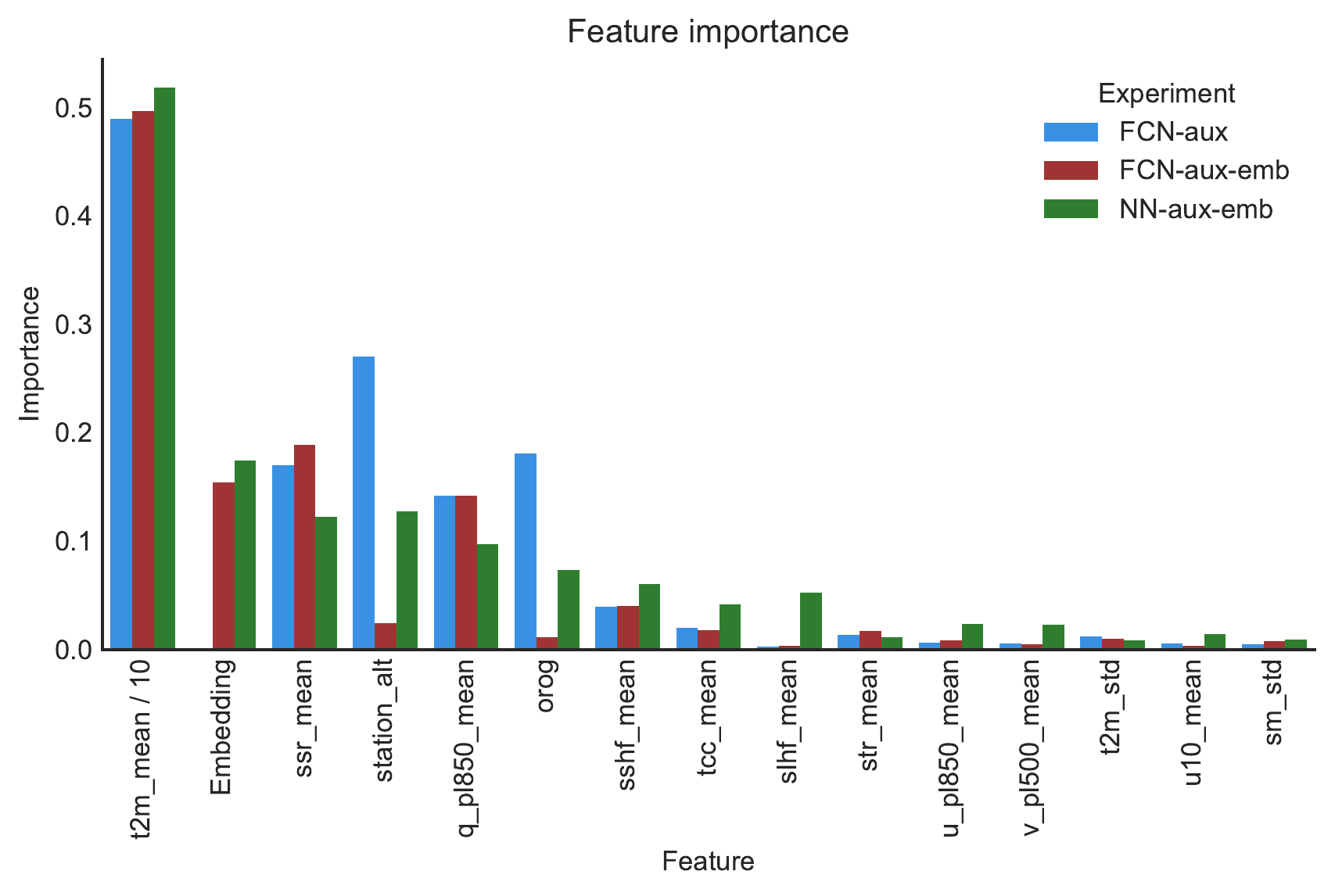}
	\caption{Feature importance for the 15 most important predictors. Note that the values for t2m\_mean are divided by ten. See Table~\ref{tab:features} for variable abbreviations and descriptions. \label{fig:feature-importance}}
\end{figure}

We picked three network setups to investigate how feature importance changes by adding station embeddings and a nonlinear layer (Figure~\ref{fig:feature-importance}). For the linear model without station embeddings (FCN-aux) station altitude and orography, the altitude of the model grid cell, are the most important predictors after the mean temperature forecast. This makes sense since our interpolation from the forecast model grid to the station does not adjust for the height of the surface station. The only other features with significant importance are the mean shortwave radiation flux and 850\,hPa specific humidity. Adding station embeddings (FCN-aux-emb) reduces the significance of the station altitude information which now seems to be encoded in the latent embedding features.  The nonlinearity added by the hidden layer in NN-aux-emb increases the sensitivity to permuting input features overall and distributes the feature importance more evenly. In particular, we note an increase in the importance of the station altitude and orography but also the sensible and latent heat flux and total cloud cover. 

The most important features, apart from the obvious mean forecast temperature and station altitude, seem to be indicative of insolation, either directly like the shortwave flux or indirectly like the 850\,hPa humidity. It is interesting that the latter seems to be picked by the algorithms as a proxy for cloud cover rather than the direct cloud cover feature. Curiously, the temperature standard deviation is not an important feature for the post-processing models. We suspect that this is a consequence of the low correlation between the raw ensemble standard deviation and the forecast error ($r=0.15$ on the test set) and the general under-dispersion (mean spread-error ratio of 0.51). The post-processing algorithms almost double the spread to achieve a spread-error ratio of 0.95. The correlation of the raw and post-processed ensemble spreads is 0.39 suggesting that the post-processing is mostly an additive correction to the ensemble spread. 

Note that this method of assessing feature importance is in principle possible for boosting- and QRF-based models. However, for the local implementations of the algorithm the importance changes from station to station, making interpretation more difficult.

\section{Discussion}\label{sec:discussion}

Here we discuss some approaches we attempted that failed to improve our results, as well as directions for future research. 

Having to describe the distribution of the target variable in parametric techniques is a non-trivial task. For temperature, a Gaussian distribution is a good approximation but for other variables, such as wind speed or precipitation, finding a distribution that fits the data is a substantial challenge \citep[e.g.,][]{TaillardatEtAl2016,BaranLerch2018}. Ideally, a machine learning algorithm would learn to predict the full probability distribution rather than distribution parameters only. One way to achieve this is to approximate the forecast distribution by a combination of uniform distributions and predicting the probability of the temperature being within pre-specified bins. Initial experiments indicate that the neural network is able to produce a good approximation of a Gaussian distribution but the skill was comparable only to the raw ensemble. This suggests that for target variables which are well approximated by a parametric distribution, utilizing these distributions is advantageous. One direction for future research is to apply this approach to more complex variables. 

Standard EMOS models are often estimated based on so-called ``rolling'' training windows with data from previous days only in order to incorporate temporal dependencies of ensemble forecast errors. For neural networks, one way to incorporate temporal dependencies is to use convolutional or recurrent neural networks \citep{Schmidhuber2015} which can process sequences as an input. In our tests, this leads to more overfitting without an improvement to the validation score. For other datasets, however, we believe that these approaches are worth revisiting. 

One popular way to combat overfitting in machine learning algorithms is data augmentation. In the example of image recognition models, the training images are randomly rotated, flipped, zoomed, etc.~to artificially increase the sample size \citep[e.g.,][]{KrizhevskyEtAl2012}. We tried a similar approach by adding random noise of a reasonable scale to the input features, but found no improvement in the validation score. A potential alternative to adding random noise might be augmenting the forecasts for a station with data from neighboring stations or grid points. 

Similarly to rolling training windows for the traditional EMOS models, we tried updating the neural network each day during the validation period with the data from the previous time step, but found no improvements. This supports our observation that rolling training windows only bring marginal improvements for the benchmark EMOS models. Such an online learning approach could be more relevant in an operational setting, however, where model versions might change frequently or it is too expensive to re-estimate the entire post-processing model every time new data becomes available. 

We have restricted the set of predictors to observation station characteristics and summary statistics (mean and standard deviation) of ensemble predictions of several weather variables. Recently, flexible distribution-to-distribution regression network models have been proposed in the machine learning literature \citep[e.g.,][]{OlivaEtAl2013,KouEtAl2018}. Adaptations of such approaches might enable the use of the entire ensemble forecast of each predictor variable as input feature. However, training of these substantially more complex models likely requires longer training periods than were possible in our study.

Another possible extension would be to post-process forecasts on the entire two-dimensional grid, rather than individual stations locations, for example by using convolutional neural networks. This adds computational complexity and probably requires more training data but could provide information about the large-scale weather patterns and help to produce spatially consistent predictions.

We have considered probabilistic forecasts of a single weather variable at a single location and look-ahead time only. However, many applications require accurate models of cross-variable, spatial and temporal dependence structures, and much recent work has been focused on multivariate post-processing methods \citep[e.g.,][]{SchefzikEtAl2013}. Extending the neural network based approaches to multivariate forecast distributions accounting for such dependencies presents a promising starting point for future research.

\section{Conclusion}\label{sec:conclusion}

In this study we demonstrated how neural networks can be used for distributional regression  post-processing of ensemble weather forecasts. Our neural network models significantly outperform state-of-the-art post-processing techniques while being computationally more efficient. The main advantages of using neural networks are the capability of capturing nonlinear relations between arbitrary predictors and distribution parameters without having to specify appropriate link functions, and the ease of adding station information in a global model by using embeddings. The network model parameters are estimated by optimizing the CRPS, a non-standard choice in the machine learning literature tailored to probabilistic forecasting. Furthermore, the rapid pace of development in the deep learning community provides flexible and efficient modeling techniques and software libraries. The presented approach can therefore be easily applied to other problems.

The building blocks of our network model architecture provide general insight into the relative importance of model properties for post-processing ensemble forecasts. Specifically, the results indicate that encoding local information is very important for providing skillful probabilistic temperature forecasts. Further, including covariate information via auxiliary variables improves the results considerably, particularly when allowing for nonlinear relations of predictors and forecast distribution parameters. Ideally, any post-processing model should thus strive to incorporate all of these aspects. 

We also showed that a trained machine learning model can be used to gain meteorological insight. In our case, it allowed us to identify the variables most important for correcting systematic temperature forecast errors of the ensemble. In this context, neural networks are somewhat interpretable and give us more information than we originally asked for. While a direct interpretation of the individual parameters of the model is intractable, this challenges the common notion of neural networks as pure black boxes. 

Because of their flexibility neural networks are ideally suited to handle the increasing amounts of model and observation data as well as the diverse requirements for correcting multifaceted aspects of systematic ensemble forecast errors. We anticipate, therefore, that they will provide a valuable addition to the modeler's toolkit for many areas of statistical post-processing and forecasting.

\section*{Acknowledgements}

The research leading to these results has been done within the subprojects A6 ``Representing forecast uncertainty using stochastic physical parameterizations'' and C7 ``Statistical postprocessing and stochastic physics for ensemble predictions'' of the Transregional Collaborative Research Center SFB/TRR 165 ``Waves to Weather'' funded by the German Research Foundation (DFG). 
Sebastian Lerch is grateful for infrastructural support by the Klaus Tschira Foundation. 
The authors thank Tilmann Gneiting, Alexander Jordan, and Maxime Taillardat for helpful discussions and for providing code.
The initial impetus for this work stems from a meeting with Kai Polsterer who presented a probabilistic neural network based approach to astrophysical image data analysis. 

\bibliographystyle{myims2}
\bibliography{bibliography}

\clearpage 

\appendix

\section{Hyperparameters for benchmark and network models}
\setcounter{table}{0}
\renewcommand{\thetable}{A\arabic{table}}
\setcounter{figure}{0}
\renewcommand{\thefigure}{A\arabic{figure}}
\setcounter{equation}{0}
\renewcommand{\theequation}{A\arabic{equation}}

\begin{table}[h]
\caption{Hyperparameters for benchmark models. AIC denotes the Akaike information criterion. \label{tab:hyper-parameters-benchmark}}
\centering
\begin{tabular}{lll}
\toprule 
Model & Parameter & Value \\
\midrule 
EMOS-gl & none & \\
EMOS-loc &  none & \\
\midrule
EMOS-loc-bst & maximum number of iterations & 1\,000 \\
	& step size & 0.05 \\
	& stopping criterion for boosting algorithm & AIC \\
\midrule
QRF &  number of trees & 1\,000 \\
	& minimum size of terminal leaves & 10 \\
	&  number of variables randomly sampled  as & 25 \\
	&  \quad   candidates at each split & \\
\bottomrule
\end{tabular}
\bigbreak
\caption{Hyperparameters for network models. Values in parentheses indicate settings for the longer training period from 2007--2015. Parameters refers to all learnable values: weights, biases and latent embedding features. An epoch refers to one pass through all training samples. Batch size refers to the number of random training samples considered per gradient update in the SGD optimization.}
\label{tab:hyper-parameters-network}
\begin{tabular}{lcccccc}
\toprule 
Model & Number of & Epochs & Learning rate & Batch size & Hidden  & Embedding \\
 & parameters & & & & nodes & size \\
\midrule 
FCN & \hfill 6   & \hfill30 (15) & \hfill 0.1 (0.1)   & \hfill4\,096 (4\,096) &    &     \\
FCN-aux & \hfill 82  & \hfill30 (10) & \hfill 0.02 (0.02)   &\hfill \,1024 (1\,024) &    &     \\
FCN-emb & \hfill 1\,084 &\hfill 30 (10) & \hfill0.02 (0.02)   & \hfill1\,024 (1\,024) &   &\hfill 2 (2)     \\
FCN-aux-emb &\hfill 1\,160 &\hfill 30 (10) &\hfill 0.02 (0.02)   &\hfill 1\,024 (1\,024) &              & \hfill2 (2)          \\
NN-aux   &\hfill 3\,326  &\hfill (10)    &\hfill (0.02)        & \hfill(1\,024)      & \hfill(64)         & \hfill(2)            \\
NN-aux-emb  & \hfill 24\,116 &\hfill 30 (10) &\hfill 0.01 (0.002)  & \hfill1\,024 (4\,096) &\hfill 50 (512)     &\hfill 2 (2)     \\
\bottomrule    
\end{tabular}
\end{table}

\section{Forecast evaluation}\label{sec:appendix-evaluation}
\setcounter{table}{0}
\renewcommand{\thetable}{B\arabic{table}}
\setcounter{figure}{0}
\renewcommand{\thefigure}{B\arabic{figure}}
\setcounter{equation}{0}
\renewcommand{\theequation}{B\arabic{equation}}

For the purpose of the present Section, denote a generic probabilistic forecast for 2-meter temperature $y_{s,t}$ at station $s$ and time $t$ by $F_{s,t}$. Note that $F_{s,t}$ may be a parametric forecast distribution represented by CDF or probability density function (PDF), an ensemble forecast $x_{s,t}^{\text{t2m},1}, \dots, x_{s,t}^{\text{t2m},50}$ or a set of quantiles. We may choose to suppress the index ${s,t}$ at times for ease of notation.

\subsection{Calibration and sharpness}

As argued by \citet{GneitingEtAl2007}, probabilistic forecasts should generally aim to maximize sharpness subject to calibration. In a nutshell, a forecast is called calibrated if the realizing observation cannot be distinguished from a random draw from the forecast distribution. Calibration thus refers to the statistical consistency between forecast distribution and observation. By contrast, sharpness is a property of the forecast only and refers to the concentration of the predictive distribution. The calibration of ensemble forecasts can be assessed via verification rank (VR) histograms summarizing the distribution of ranks of the observation $y_{s,t}$ when it is pooled with the ensemble forecast \citep{Hamill2001,GneitingEtAl2007,Wilks2006}. For continuous forecast distributions histograms of the PIT $F_{s,t}(y_{s,t})$ provide analogs of verification rank histograms. Calibrated forecasts result in uniform VR and PIT histograms, and deviations from uniformity indicate specific systematic errors such as biases or an under-representation of the forecast uncertainty. 

\subsection{Proper scoring rules}

For comparative model assessment, proper scoring rules allow simultaneous evaluation of calibration and sharpness \citep{GneitingRaftery2007}. A scoring rule $S(F,y)$ assigns a numerical score to a pair of probabilistic forecast $F$ and corresponding realizing observation $y$, and is called proper relative to a class of forecast distributions $\mathcal{F}$ if
\[
\mathbb{E}_{Y\sim G}S(G,Y) \leq \mathbb{E}_{Y\sim G}S(F,Y) \text{ for all } F,G\in\mathcal{F},
\]
i.e., if the expected score is optimized if the true distribution of the observation is issued as forecast. Here, scoring rules are considered to be negatively oriented with smaller scores indicating better forecasts

Popular examples of proper scoring rule include the logarithmic score \citep[LogS;][]{Good1952}
\begin{equation*}
\textnormal{LogS}(F,y) = -\log(f(y)),
\end{equation*}
where $y$ denotes the observation and $f$ denotes the PDF of the forecast distribution, and the continuous ranked probability score \citep[CRPS;][]{MathesonWinkler1976} 
\begin{equation}\label{eq:crps}
\textnormal{CRPS}(F,y) = \int_{-\infty}^{\infty} (F(z) - \mathbbm{1}(y \leq z))^2 \textnormal{d}z,
\end{equation}
where $F$ denotes the CDF of the forecast distribution with finite first moment and $\mathbbm{1}(y \leq z)$ is an indicator function that is 1 if $y \leq z$ and 0 otherwise. The integral in \eqref{eq:crps} can be computed analytically for ensemble forecasts and a variety of continuous forecast distributions \citep[see, e.g.,][]{JordanEtAl2017}. Specifically, the CRPS of a Gaussian distribution with mean value $\mu$ and standard deviation $\sigma$ can be computed as
\begin{equation}\label{eq:crps-normal}
\textnormal{CRPS}(F_{\mu, \sigma}, y) = \sigma\cdot\left( \tfrac{y - \mu}{\sigma}\left(2\Phi(\tfrac{y - \mu}{\sigma})-1\right) + 2\varphi(\tfrac{y - \mu}{\sigma})  - \frac{1}{\sqrt{\pi}} \right),
\end{equation}
where $\Phi$ and $\varphi$ denote CDF and PDF of a standard Gaussian distribution, respectively \citep{GneitingEtAl2005}.

Apart from forecast evaluation, proper scoring rules can also be used for parameter estimation. Following the generic optimum score estimation framework of \citet[Section 9.1]{GneitingRaftery2007}, the parameters of a forecast distribution are determined by optimizing the value of a proper scoring rule, on average over a training sample. Optimum score estimation based on the LogS then corresponds to classical maximum likelihood estimation, whereas optimum score estimation based on the CRPS is often employed as a more robust alternative in meteorological applications. Analytical closed-form solutions of the CRPS, for example for a Gaussian distribution in \eqref{eq:crps-normal}, allow for computing analytical gradient functions that can be leveraged in numerical optimization, see \citet{JordanEtAl2017} for details.

In practical applications, scoring rules are usually computed as averages over stations and/or time periods. To assess the relative improvement over a reference forecast $F_{\textnormal{ref}}$, we further introduce the continuous ranked probability skill score 
\[
\textnormal{CRPSS}(F, y) = 1-\frac{\textnormal{CRPSS}(F,y)}{\textnormal{CRPSS}(F_{\textnormal{ref}}, y)}
\]
which is positively oriented and can be interpreted as relative improvement over the reference. The CRPSS is usually computed as skill score of CRPS averages. 

\subsection{Statistical tests of equal predictive performance}\label{sec:appendix-tests}

Formal statistical tests of equal forecast performance for assessing statistical significance of score differences have been widely used in the economic literature. Consider two forecasts $F^1$ and $F^2$, with corresponding mean scores $\bar{S}(F^i) = \frac{1}{n} \sum_{j=1}^{n} S(F^i_j,y_j)$ for $i = 1,2$ over a test $j=1,\dots,n$, where we assume that the forecast $F^i_j$ was issued $k$ time steps before the observation $y_j$ was recorded. \citet{DieboldMariano1995} propose the test statistic
\[
t_n = \sqrt{n}\frac{\bar{S}(F^1) - \bar{S}(F^2)}{\hat{\sigma}_n},
\] 
where $\hat{\sigma}_n$ is an estimator of the asymptotic standard deviation of the score difference of $F^1$ and $F^2$. Under standard regularity conditions, $t_n$ asymptotically follows a standard normal distribution under the null hypothesis of equal predictive performance of $F^1$ and $F^2$. Thereby, negative values of $t_n$ indicate superior predictive performance of $F^1$, whereas positive values indicate superior performance of $F^2$. To account for temporal dependencies of score differences, we use the square root of the sample auto-covariance up to lag $k-1$ as estimator $\hat{\sigma}_n$ following \citet{DieboldMariano1995}. We employ Diebold-Mariano tests on an observation station-level, i.e., the mean CRPS values are determined by averaging over all scores at the specific station $s_0\in\{1,\dots,S\}$ of interest,
\[
\overline{\text{CRPS}}(F^i_{s_0}) = \frac{1}{T} \sum_{t=1}^{T} F^i_{{s_0},t},
\]  
where $t = 1,\dots,T$ denotes days in the evaluation period. 

Compared to previous uses of Diebold-Mariano tests in post-processing applications \citep[e.g.,][]{BaranLerch2016} we further account for spatial dependencies of score differences at the different stations. Following suggestions of \citet{Wilks2016} we apply a \citet{BenjaminiHochberg1995} procedure to control the false discovery rate at level $\alpha$. In a nutshell, the algorithm requires a higher standard in order to reject a local null hypothesis of equal predictive performance by selecting a threshold $p$-value $p^\ast$ based on the set of ordered local $p$-values $p_{(1)},\dots,p_{(S)}$. Particularly,  $p^\ast$ is the largest $p_{(i)}$ that is not larger than $i/S\cdot\alpha$, where $S$ is the number of tests, i.e., the number of stations in the evaluation set.

\clearpage 

\section{Additional results}
\setcounter{table}{0}
\renewcommand{\thetable}{C\arabic{table}}
\setcounter{figure}{0}
\renewcommand{\thefigure}{C\arabic{figure}}
\setcounter{equation}{0}
\renewcommand{\theequation}{C\arabic{equation}}

\subsection{Calibration assessment}

\begin{figure}[!htb]
\includegraphics[width=0.95\textwidth]{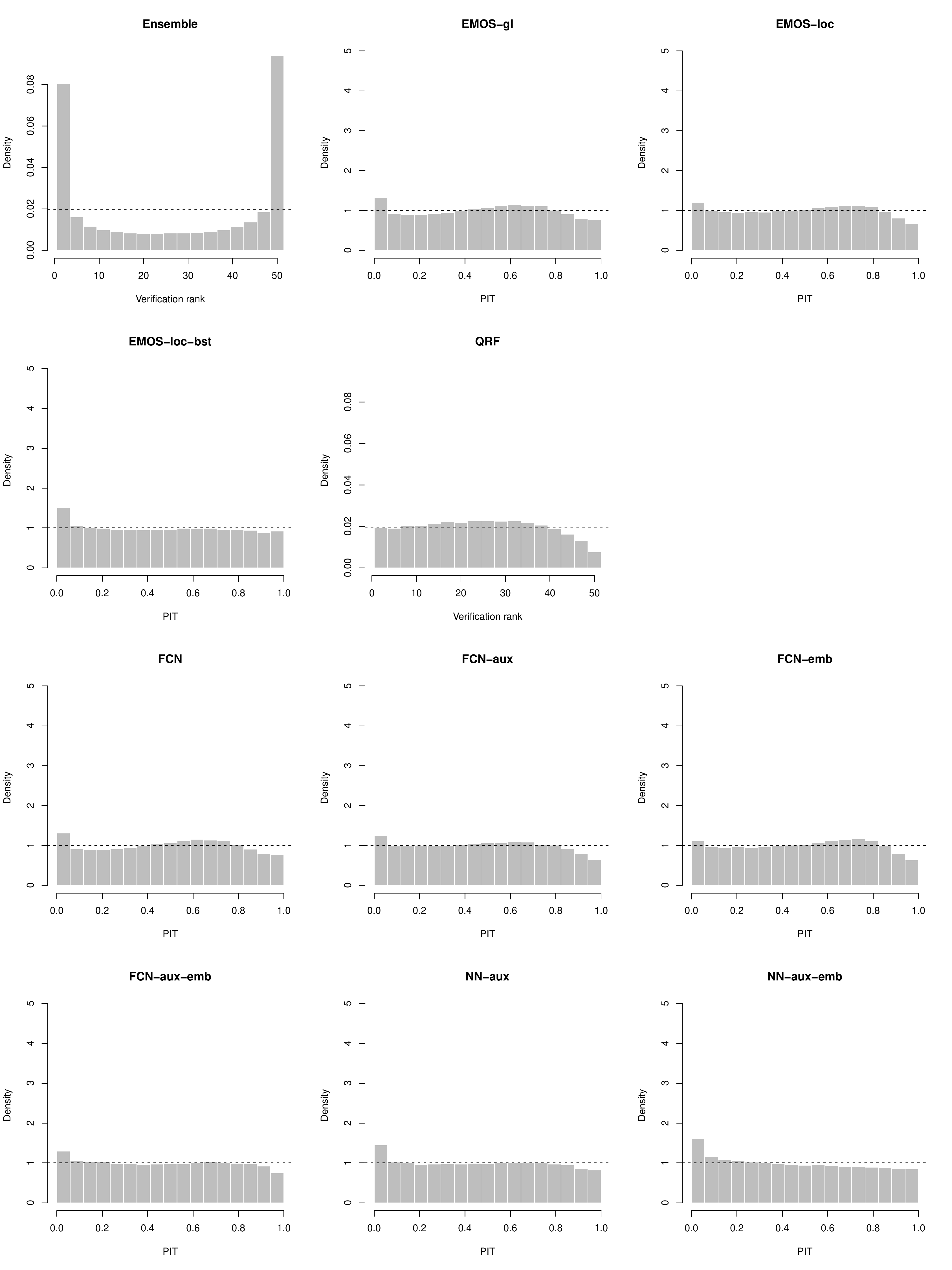}
\caption{Verification rank and PIT histograms for raw and post-processed ensemble forecasts based on models estimated using data from 2015, aggregated over all forecast cases during the evaluation period in calendar year 2016. \label{fig:pits-2015}}
\end{figure}

\begin{figure}[!htb]
\includegraphics[width=0.95\textwidth]{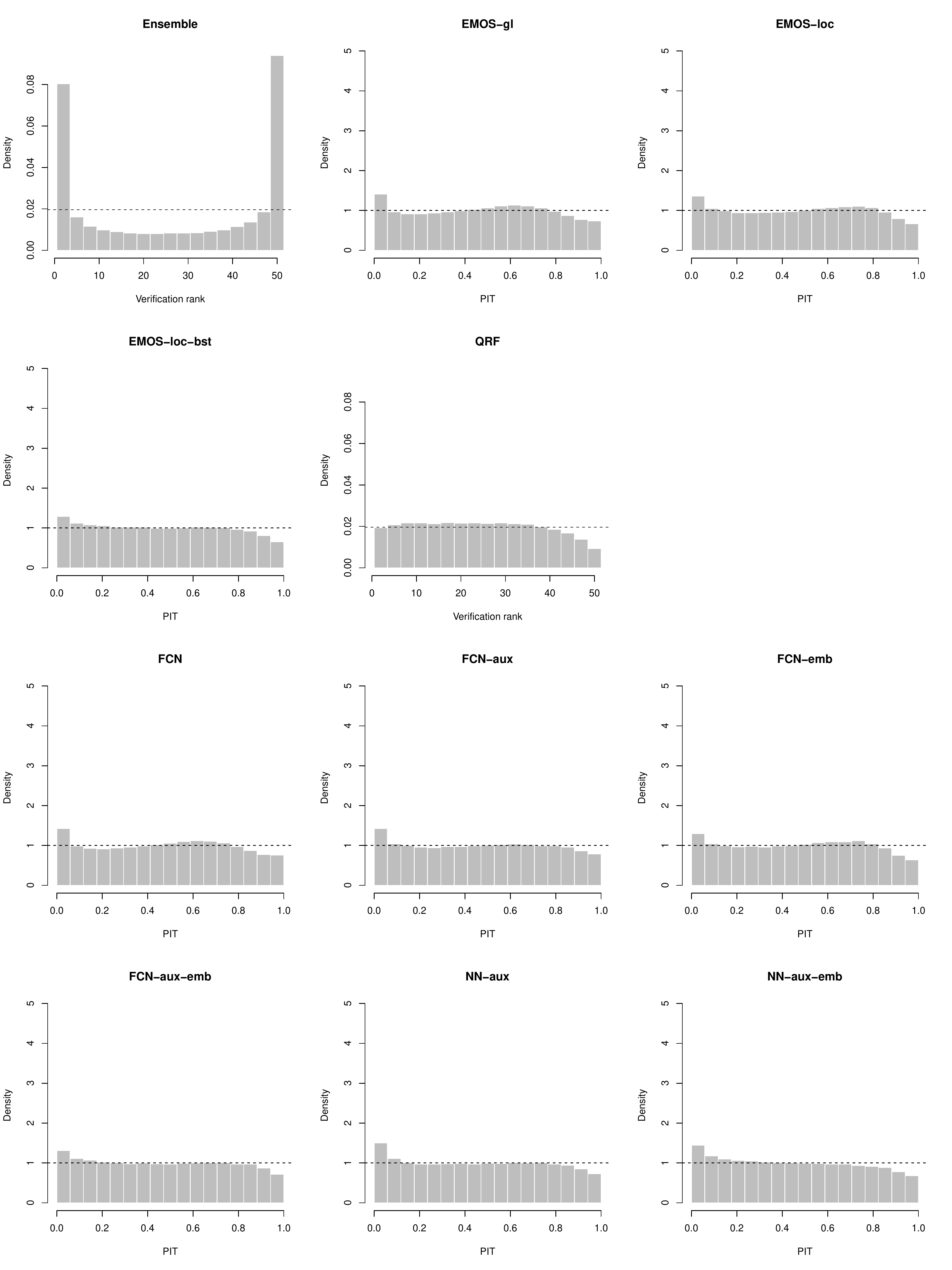}
\caption{Verification rank and PIT histograms for raw and post-processed ensemble forecasts based on models estimated using data from 2007--2015, aggregated over all forecast cases during the evaluation period in calendar year 2016.. \label{fig:pits-2007-15}}
\end{figure}

\clearpage 

\subsection{CRPSS results for alternative benchmark models}\label{sec:appendix-results-crpss}

\begin{figure}[!htb]
\centering
CRPSS relative to EMOS-loc-boost \\
\includegraphics[width = 0.71\textwidth]{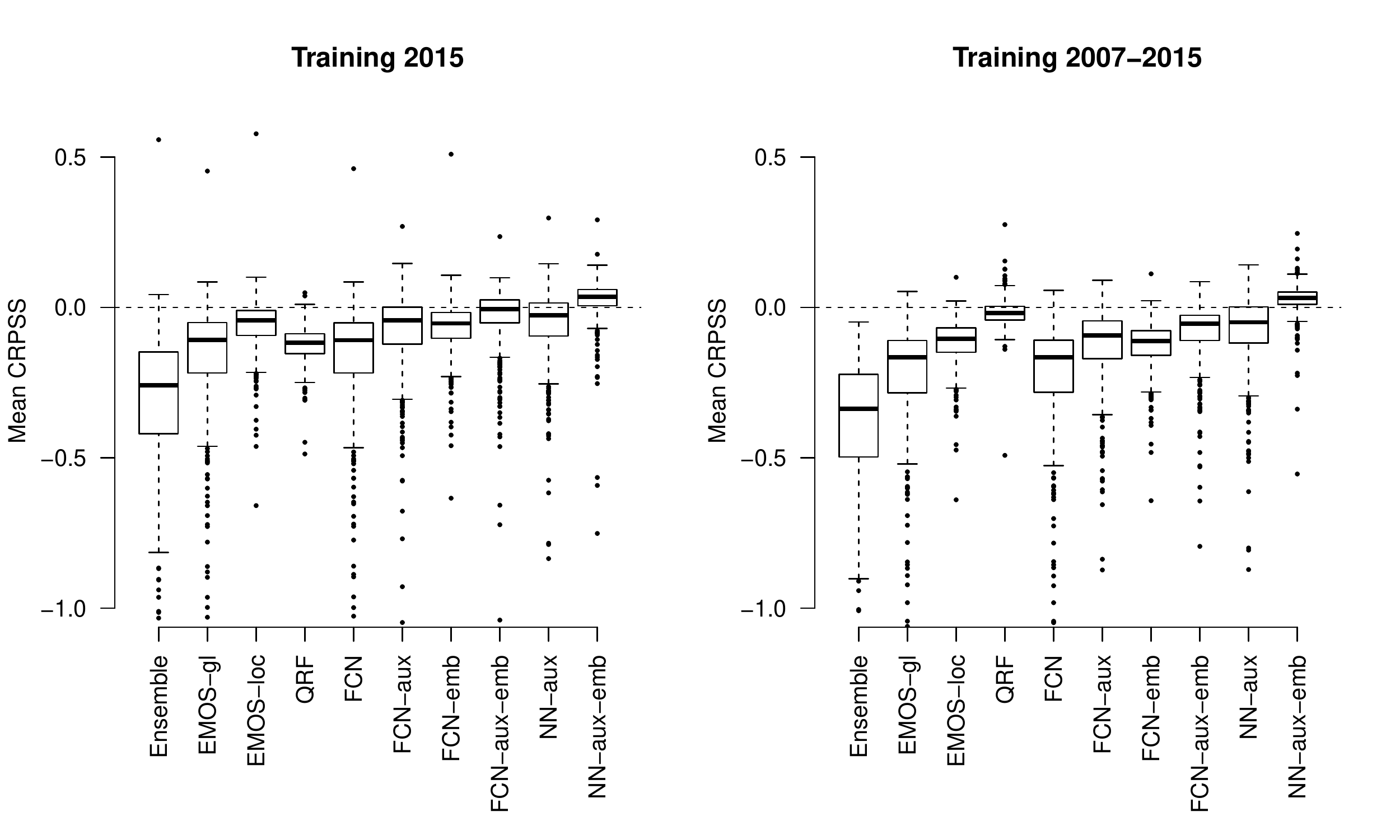} \\
\smallbreak
CRPSS relative to QRF \\
\includegraphics[width = 0.71\textwidth]{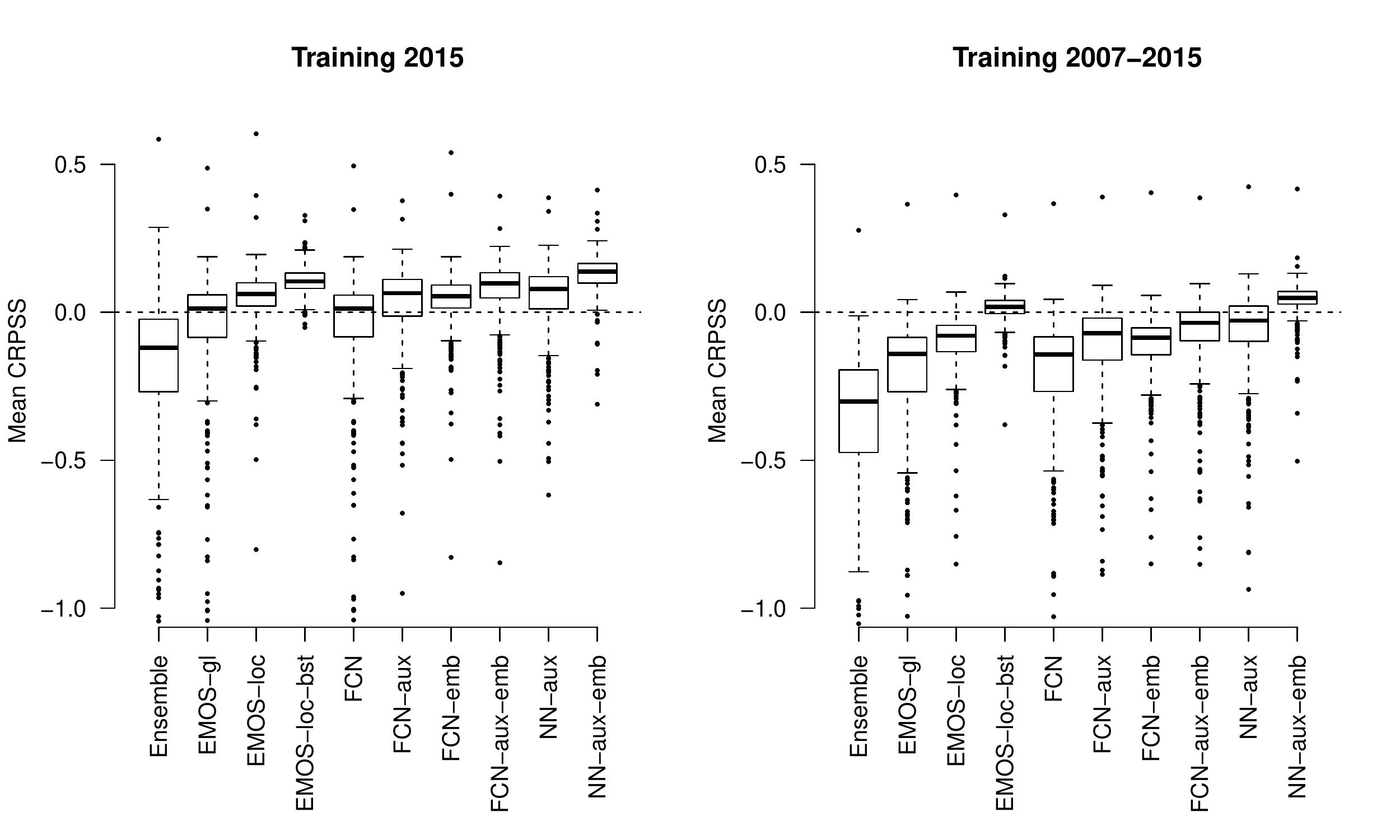} \\
\smallbreak
CRPSS relative to NN-aux-emb \\
\includegraphics[width = 0.71\textwidth]{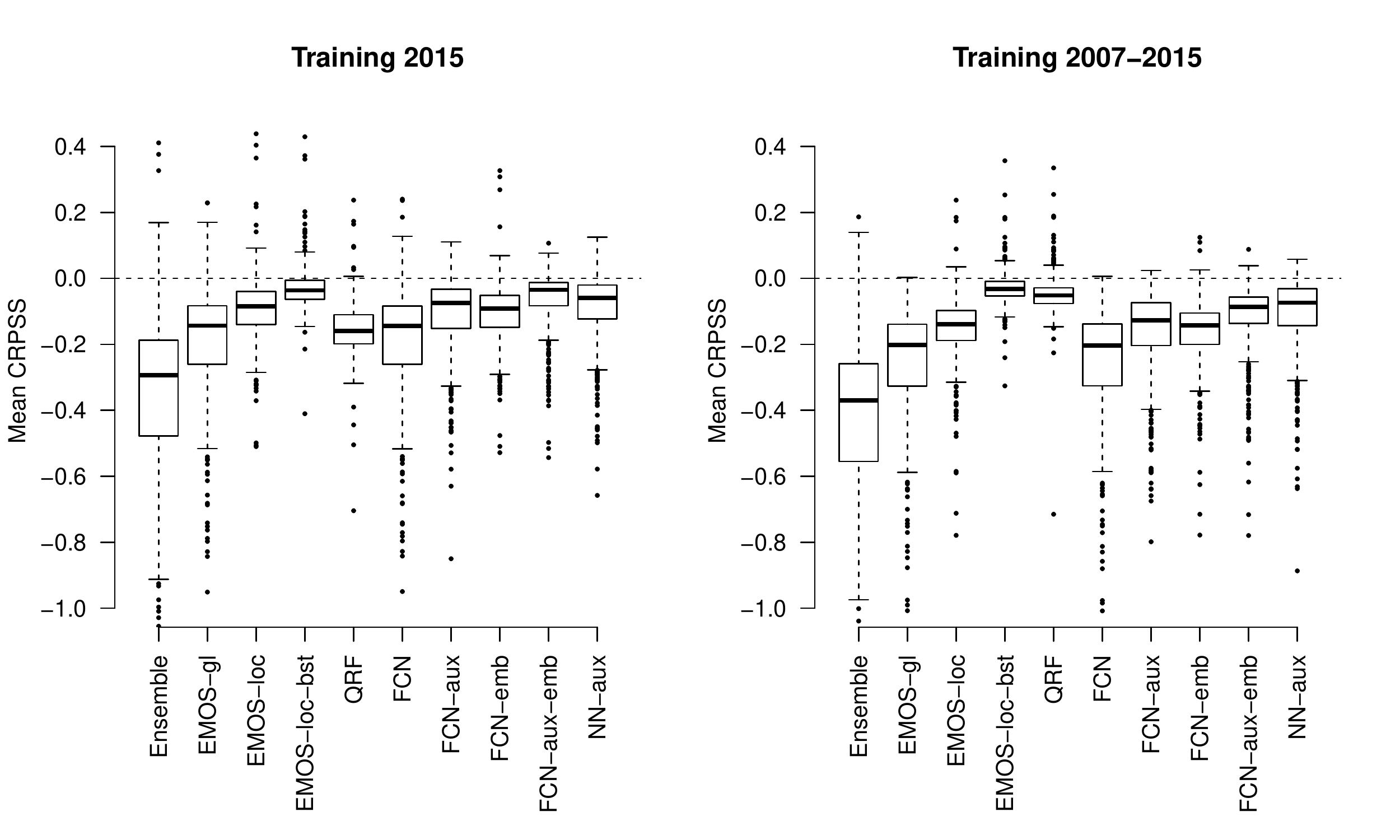} \\
\caption{As Figure \ref{fig:crpss}, but with different reference models. \label{fig:crpss-appendix}}
\end{figure}

\clearpage 

\subsection{Details on computational aspects}\label{sec:appendix-results-computation}

Table \ref{tab:computation} shows computation times required for training the different post-processing models for both training sets. As noted before, the computation times are not directly comparable due to implementations in different programming languages and hardware environments.  The computation times for the benchmark models, implemented in \texttt{R} using the \texttt{crch} \citep{MessnerEtAl2016}, \texttt{quantregForest} \citep{quantregForest} and \texttt{scoringRules} \citep{JordanEtAl2017} packages, were obtained on a standard laptop computer, whereas the network models were implemented with the \texttt{Python} libraries Keras \citep{Keras} and TensorFlow \citep{Tensorflow}, and run on a single GPU (Nvidia Tesla K20). Computation times on a regular CPU are roughly 6 times longer for the most complex networks. For the simple networks the difference is negligible. Note that the inference time, i.e., the time to make a prediction after the model has been trained, is on the order of a few seconds for all models. Further, note that all computation times reported here are substantially lower compared to the computational costs of generating the raw ensemble forecast.

\begin{table}[!hbt]
\caption{Computation times (in minutes) for estimating post-processing models with the two training sets and computing out-of-sample forecasts for the evaluation period.\label{tab:computation}}
\centering
\begin{tabular}{lrr}
\toprule
Model & \multicolumn{2}{c}{Computation time (min)} \\
 & \multicolumn{2}{c}{with training data from} \\[0.15cm]
 & 2015 & 2007--2015 \\
\midrule
\multicolumn{3}{l}{\textit{Benchmark models}} \\
\midrule 
 EMOS-gl & $< 1$ & $< 1$ \\
 EMOS-loc & $< 1$ & 1 \\
 EMOS-loc-bst & 14 & 48 \\
 QRF & 8 & 430 \\
\midrule
\multicolumn{3}{l}{\textit{Network models}} \\
\midrule 
FCN  & $< 1$ & 1 \\
FCN-aux  & $< 1$ & 2 \\
FCN-emb & $< 1$ & 3 \\
FCN-aux-emb & $< 1$ & 3 \\
NN-aux & 4 & 25 \\
NN-aux-emb & 9 & 16 \\
\bottomrule
\end{tabular}
\end{table}

\clearpage 

\newgeometry{left=24mm,right=24mm,top=25mm,bottom=10mm}

\subsection{Statistical significance of score differences}\label{sec:appendix-results-significance}

Pair-wise one-sided Diebold-Mariano tests are applied to all possible comparisons of forecast models at each of the 499 stations individually. To account for multiple hypothesis testing and spatial correlations of score differences, we apply a Benjamini-Hochberg procedure to the corresponding $p$-values when aggregating the results by determining the ratio of stations with significant score differences, see Appendix \ref{sec:appendix-tests} for details. 

Table \ref{tab:dmtests} summarizes pair-wise Diebold-Mariano tests by showing the ratio of stations with statistically significant CRPS differences after applying a Benjamini-Hochberg procedure for a nominal level of $\alpha = 0.05$. Generally, the results indicate large numbers of stations with significant differences of the network models when compared to standard EMOS approaches. NN-aux-emb shows the highest ratios of significant score differences over any competitor, and is significantly outperformed at very few station and only by the best-performing alternatives. 

\begin{table}[hb]
	\caption{Ratio of stations (in \%) where pair-wise Diebold-Mariano tests indicate statistically significant CRPS differences after applying a Benjamini-Hochberg procedure to account for multiple testing for a nominal level of $\alpha=0.05$ of the corresponding one-sided tests. The $(i,j)$-entry in the $i$-th row and $j$-th column indicates the ratio of stations where the null hypothesis of equal predictive performance of the corresponding one-sided Diebold-Mariano test is rejected in favor of the model in the $i$-th row when compared to the model in the $j$-th column. The remainder of the sum of $(i,j)$- and $(j,i)$-entry to 100\% is the ratio of stations where the score differences are not significant.  
	\label{tab:dmtests}}
	\centering
	\bigbreak 
	Training with 2015 data
	\bigbreak 
	\small 
	\begin{tabular}{rccccccccccc}
		\toprule
		& {\footnotesize{Ens.}} & {\footnotesize{EMOS}}  & {\footnotesize{EMOS}} & {\footnotesize{EMOS}} &  {\footnotesize{QRF}} & {\footnotesize{FCN}} & {\footnotesize{FCN}} & {\footnotesize{FCN}} & {\footnotesize{FCN}} & {\footnotesize{NN}} & {\footnotesize{NN}} \\ 
		&		&	{\footnotesize{-gl}} & {\footnotesize{-loc}} & {\footnotesize{-loc-bst}}  &  & & {\footnotesize{-aux}} & {\footnotesize{-emb}}   & {\footnotesize{-aux-emb}} & {\footnotesize{-aux}} & {\footnotesize{-aux-emb}} \\	
		\midrule
		{\footnotesize{Ens.}}&  & \hphantom{0}0.6 & \hphantom{0}0.0 & \hphantom{0}0.0 & \hphantom{0}0.6 & \hphantom{0}0.6 & \hphantom{0}0.8 & \hphantom{0}0.0 & \hphantom{0}0.8 & \hphantom{0}0.8 & \hphantom{0}0.6 \\ 
		{\footnotesize{EMOS-gl}} & 83.2 &  & \hphantom{0}0.2 & \hphantom{0}0.0 & 10.4 & 10.2 & \hphantom{0}3.0 & \hphantom{0}0.2 & \hphantom{0}0.6 & \hphantom{0}2.0 & \hphantom{0}0.2 \\ 
		{\footnotesize{EMOS-loc}}  & 96.2 & 71.3 &  & \hphantom{0}0.0 & 50.5 & 71.9 & 17.4 & 24.8 & \hphantom{0}5.2 & \hphantom{0}9.6 & \hphantom{0}1.4 \\ 
		{\footnotesize{EMOS-loc-bst}} & 93.8 & 72.7 & 40.5 &  & 89.8 & 74.3 & 41.7 & 49.1 & 21.0 & 30.5 & \hphantom{0}2.0 \\ 
		{\footnotesize{QRF}} & 54.7 & 22.0 & \hphantom{0}3.6 & \hphantom{0}0.0 &  & 22.4 & \hphantom{0}8.0 & \hphantom{0}3.6 & \hphantom{0}3.4 & \hphantom{0}5.2 & \hphantom{0}0.2 \\ 
		{\footnotesize{FCN}} & 83.0 & \hphantom{0}7.4 & \hphantom{0}0.2 & \hphantom{0}0.0 & 10.4 &  & \hphantom{0}3.0 & \hphantom{0}0.2 & \hphantom{0}0.6 & \hphantom{0}2.0 & \hphantom{0}0.2 \\ 
		{\footnotesize{FCN-aux}} & 83.2 & 60.3 & 17.2 & \hphantom{0}1.8 & 47.5 & 62.3 &  & 19.0 & \hphantom{0}1.0 & \hphantom{0}0.4 & \hphantom{0}0.2 \\ 
		{\footnotesize{FCN-emb}} & 89.4 & 67.1 & \hphantom{0}1.0 & \hphantom{0}0.0 & 44.1 & 68.1 & 11.4 &  & \hphantom{0}0.8 & \hphantom{0}6.4 & \hphantom{0}0.6 \\ 
		{\footnotesize{FCN-aux-emb}}  & 86.6 & 78.8 & 53.1 & \hphantom{0}7.6 & 69.1 & 79.6 & 55.1 & 58.5 &  & 27.1 & \hphantom{0}0.2 \\ 
		{\footnotesize{NN-aux}} & 87.2 & 69.5 & 25.9 & \hphantom{0}2.0 & 57.5 & 70.7 & 22.8 & 30.9 & \hphantom{0}8.0 &  & \hphantom{0}0.4 \\   
		{\footnotesize{NN-aux-emb}}  & 93.6 & 89.4 & 67.1 & 30.3 & 92.2 & 90.2 & 67.3 & 72.7 & 43.5 & 64.9 &  \\ 
		\bottomrule
	\end{tabular}
	\bigbreak\bigbreak
	{\normalsize{Training with 2007-2015 data}}
	\bigbreak
	\begin{tabular}{rccccccccccc}
		\toprule
		& {\footnotesize{Ens.}} & {\footnotesize{EMOS}}  & {\footnotesize{EMOS}} & {\footnotesize{EMOS}} &  {\footnotesize{QRF}} & {\footnotesize{FCN}} & {\footnotesize{FCN}} & {\footnotesize{FCN}} & {\footnotesize{FCN}} & {\footnotesize{NN}} & {\footnotesize{NN}} \\ 
		&		&	{\footnotesize{-gl}} & {\footnotesize{-loc}} & {\footnotesize{-loc-bst}}  &  & & {\footnotesize{-aux}} & {\footnotesize{-emb}}   & {\footnotesize{-aux-emb}} & {\footnotesize{-aux}} & {\footnotesize{-aux-emb}} \\	
		\midrule
		{\footnotesize{Ens.}} &  & \hphantom{0}0.6 & \hphantom{0}0.0 & \hphantom{0}0.0 & \hphantom{0}0.0 & \hphantom{0}0.6 & \hphantom{0}0.8 & \hphantom{0}0.0 & \hphantom{0}0.8 & \hphantom{0}0.8 & \hphantom{0}0.0 \\ 
		{\footnotesize{EMOS-gl}} & 86.8 &  & \hphantom{0}0.2 & \hphantom{0}0.0 & \hphantom{0}0.2 & \hphantom{0}2.6 & \hphantom{0}3.0 & \hphantom{0}0.2 & \hphantom{0}0.6 & \hphantom{0}0.2 & \hphantom{0}0.0 \\ 
		{\footnotesize{EMOS-loc}} & 98.8 & 72.7 &  & \hphantom{0}0.0 & \hphantom{0}0.2 & 71.7 & 17.2 & 17.4 & \hphantom{0}3.6 & \hphantom{0}6.8 & \hphantom{0}0.6 \\ 
		{\footnotesize{EMOS-loc-bst}}& 99.4 & 98.0 & 91.4 &  & 21.0 & 97.8 & 82.0 & 94.2 & 70.3 & 49.7 & \hphantom{0}1.4 \\ 
		{\footnotesize{QRF}} & 98.6 & 94.2 & 79.2 & \hphantom{0}1.4 &  & 94.2 & 57.7 & 84.4 & 38.1 & 33.5 & \hphantom{0}1.2 \\ 
		{\footnotesize{FCN}} & 87.8 & 11.0 & \hphantom{0}0.2 & \hphantom{0}0.0 & \hphantom{0}0.2 &  & \hphantom{0}3.2 & \hphantom{0}0.2 & \hphantom{0}0.6 & \hphantom{0}0.2 & \hphantom{0}0.0 \\ 
		{\footnotesize{FCN-aux}}& 87.6 & 65.5 & 24.2 & \hphantom{0}0.0 & \hphantom{0}0.4 & 65.5 &  & 26.7 & \hphantom{0}0.8 & \hphantom{0}1.4 & \hphantom{0}0.0 \\ 
		{\footnotesize{FCN-emb}} & 93.4 & 71.3 & \hphantom{0}0.0 & \hphantom{0}0.0 & 0.2 & 70.5 & 12.0 &  & \hphantom{0}1.2 & \hphantom{0}4.6 & \hphantom{0}0.0 \\ 
		{\footnotesize{FCN-aux-emb}} & 91.2 & 82.8 & 60.3 & \hphantom{0}0.0 & \hphantom{0}0.6 & 81.8 & 58.1 & 64.1 &  & 16.4 & \hphantom{0}0.0 \\ 
		{\footnotesize{NN-aux}} & 95.6 & 84.8 & 54.5 & \hphantom{0}1.4 & 9.8 & 84.8 & 72.9 & 58.5 & 34.5 &  & \hphantom{0}0.0 \\ 
		{\footnotesize{NN-aux-emb}} & 98.8 & 97.8 & 95.2 & 29.9 & 52.9 & 97.6 & 92.0 & 96.0 & 91.0 & 74.5 &  \\ 
		\bottomrule
	\end{tabular}
\end{table}

\restoregeometry

\end{document}